%% file: main.tex
\documentclass[10pt,twocolumn,letterpaper]{article}

\usepackage[pagenumbers]{iccv}              

\definecolor{cvprblue}{rgb}{0.21,0.49,0.74}
\usepackage[pagebackref,breaklinks,colorlinks,allcolors=cvprblue]{hyperref}
\usepackage{float}
\usepackage{amsmath}
\usepackage{multicol}
\usepackage{graphicx}
\usepackage{siunitx}
\usepackage{tikz}
\usepackage{multirow}
\usepackage{adjustbox}

\definecolor{best}{rgb}{1.000, 0.965, 0.698}
\definecolor{second_best}{rgb}{1.000, 0.800, 0.600}
\definecolor{third_best}{rgb}{1.000, 0.600, 0.600}

\usepackage{fontawesome}
\newcommand{\greenup}{\textcolor{green}{\faArrowUp}}
\newcommand{\reddown}{\textcolor{red}{\faArrowDown}}

\def\paperID{8658} 
\def\confName{ICCV}
\def\confYear{2025}

\title{Lightweight Gradient-Aware Upscaling of 3D Gaussian Splatting Images}

\input{authors}

\begin{document}
\maketitle
\begin{abstract}
We introduce an image upscaling technique tailored for 3D Gaussian Splatting (3DGS) on lightweight GPUs. 
Compared to 3DGS, it achieves significantly higher rendering speeds and reduces commonly observed reconstruction artifacts.
Our technique upscales low-resolution 3DGS renderings with a marginal increase in cost by directly leveraging the analytical image gradients of Gaussians for gradient-based bicubic spline interpolation.
The technique is agnostic to the specific 3DGS implementation, achieving 
novel view synthesis at rates 3×–4× higher than the baseline implementation.
Through extensive experiments on multiple datasets, we showcase the performance improvements and high reconstruction fidelity attainable with gradient-aware upscaling of 3DGS images.
We further demonstrate the integration of gradient-aware upscaling into the gradient-based optimization of a 3DGS model and analyze its effects on reconstruction quality and performance. 
\end{abstract}

\section{Introduction}

\begin{figure}[t!]
    \centering
    \includegraphics[width=\linewidth]{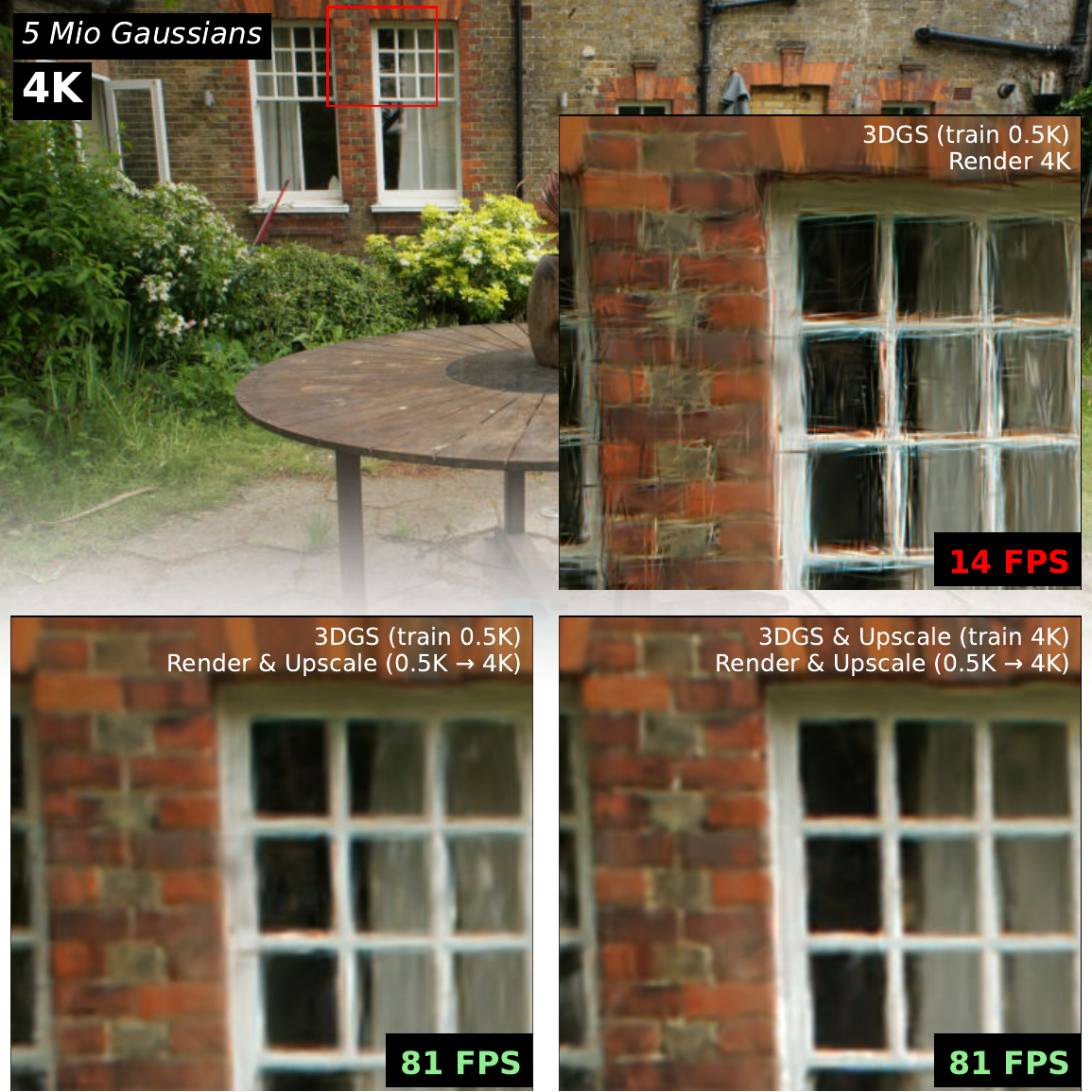}
    \caption{Bottom-left: When used as image upscaler during rendering, our method achieves higher quality and faster rendering than 3DGS (top-right). With upscaling embedded into training (bottom-right), the quality is further improved.}
    \label{fig:teaser}
\end{figure}

Despite continuous improvements of 3D Gaussian Splatting (3DGS) \cite{kerbl_3d_2023} in reconstruction~\cite{Yu2024MipSplatting,taming3dgs} and rendering performance~\cite{Niedermayr_2024_CVPR},
even high-end GPUs struggle to maintain a stable 30 FPS when rendering to high-resolution displays. 
This limitation is even more critical in VR applications, where two high-resolution images must be rendered simultaneously 
to maintain a smooth and immersive experience.
When 3DGS models are accessed via mobile devices with high-resolution screens but lightweight GPUs~\cite{niedermayr24cinematic}, rendering performance often drops to an unacceptably low level.

\begin{figure}[h!]
    \centering
    \includegraphics[width=\linewidth]{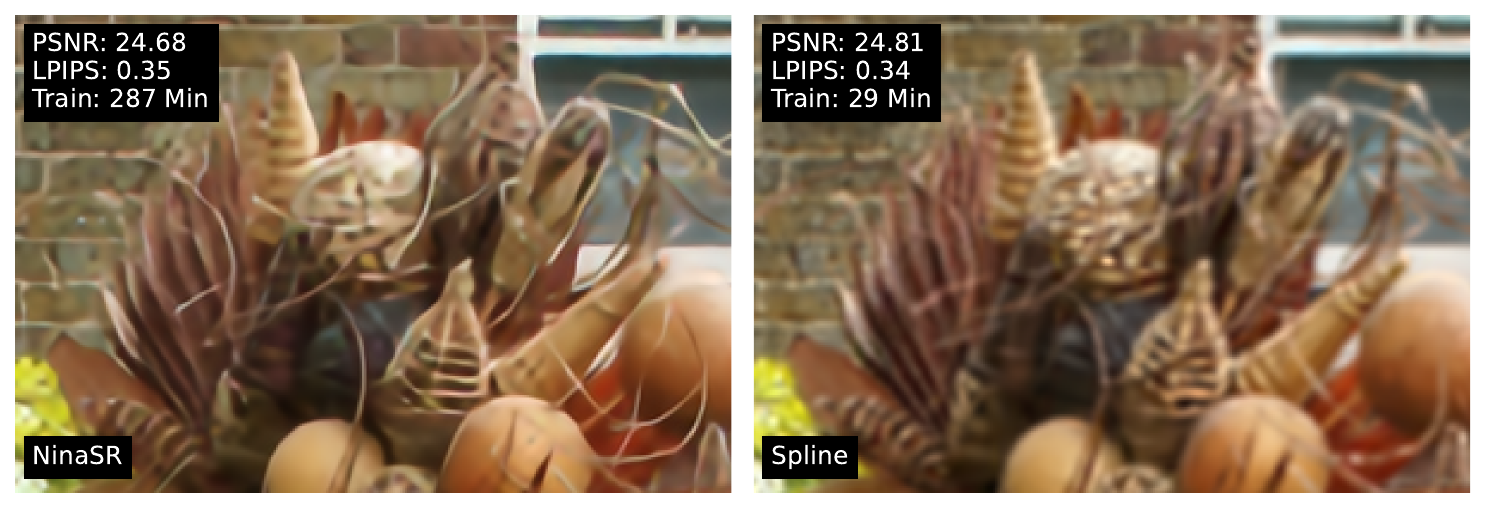}
    \caption{4x DL-based vs. Spline-based (ours) upscaling.}
    \label{fig:nina_vs_spline}
    \vspace{-.1cm}
\end{figure}

A common strategy to 
render to higher-resolution displays without increasing the rasterization load is image upscaling.
Image upscaling methods such as Bicubic and Lanczos interpolation are widely used~\cite{amd_fsr} but often reduce visual quality by introducing noticeable artifacts.
Deep learning (DL)-based upscalers offer superior quality by 
enhancing image details, but they also have significant drawbacks. 
High-quality approaches like NinaSR \cite{ninasr}, SRGS \cite{feng2024srgs} or SwinIR \cite{SwinIR} are not fast enough for real-time applications, while solutions such as NVIDIA DLSS~\cite{nvidia_dlss} require specialized hardware and rely on the use of additional image information, such as accurate depth buffers and motion vectors, which are unavailable for 3DGS. 
Furthermore, when using DL-based upscalers in VR, users often report issues due to temporal inconsistencies and feature popping. 
Although such upscalers produce sharp images, the results often appear unnatural due to hallucinated artifacts, and PSNR scores tend to decrease (see Fig.~\ref{fig:nina_vs_spline}).


We propose a computationally lightweight image upscaling technique specifically tailored for 3DGS to address these limitations. 
Innovation comes from uniquely leveraging Gaussian primitives' gradients, introducing a new way of thinking about upscaling: not using fixed kernels as in most classical methods, not requiring heavy pre-training and high upscaling times as modern neural approaches, and delivering faster upscaling with superior image quality than highly optimized GPU-based upscalers.  

Compared to 3DGS, we achieve significantly improved frame rates at high fidelity. The quality-improved variant of 3DGS, Mip-Splatting \cite{Yu2024MipSplatting}, shows limited quality reconstruction of regions sparsely viewed in the test set and 
further increases the rasterization workload (see supplemental material).

We further demonstrate that the upscaling process can be integrated into the gradient-based optimization of a 3DGS model. This allows optimization on low-resolution renderings that are subsequently upscaled, further increasing reconstruction accuracy. 
By enhancing rendering efficiency without relying on deep learning inference or additional scene information, high-quality and temporally stable upscaling for 3DGS can be performed on lightweight GPUs with limited rasterization capabilities.

Our specific contributions are:

\begin{itemize}
    \item Gradient-aware Gaussian upscaling: We derive a bicubic spline-based image interpolation technique that directly incorporates the analytical gradients of the Gaussians used to generate the image.  
    \item Analytical gradient backpropagation: To optimize the 3DGS model for upscaled rendering, we demonstrate the backpropagation of analytical gradients during 3DGS model optimization. 
    \item High-speed, high-quality upscaling: Through extensive experiments, we demonstrate the performance improvements and high reconstruction fidelity attainable with gradient-aware upscaling of 3D Gaussian splatting, as compared to classical bicubic interpolation, Lanczos resampling, and DL-based upscaling. 
\end{itemize}


\section{Related Work}

\paragraph{Image upscaling.} Image upscaling, also known as super-resolution (SR) rendering, is a technique that generates high-resolution images from rendered lower-resolution inputs.
It is especially useful in VR/AR environments and gaming on lightweight  devices, where it enables high-quality visuals at interactive framerates.
Super-resolution algorithms range from traditional interpolation techniques, such as bilinear or bicubic interpolation, to more advanced upscalers, like Lanczos filtering and deep learning-based upscaling techniques.

Compared to bilinear interpolation, which uses a weighted average of surrounding pixels, bicubic interpolation fits a cubic polynomial to smooth the transition between pixels, with the tendency to blur out the image at larger scales.
Lanczos resampling, which is used in AMD's FidelityFX Super Resolution (FSR)~\cite{amd_fsr} technology, uses a finite windowed sinc function to better preserve high-frequency details, yet at a significantly higher computational cost. 

DL-based super-resolution using convolutional neural networks (CNNs) or Generative Adversarial Networks (GANs), trained on low- and high-resolution image pairs, can synthesize realistic details, sharp edges, and textures, making the upscaled image nearly indistinguishable from a true high-resolution rendering \cite{dong2015image,dong2016accelerating,lai2017deep,lim2017edsr,ledig2017photo,wang2018esrgan,wang2021realesrgan,torchsr,ninasr}.
For real-time applications, the most suitable approaches are DLSS \cite{nvidia_dlss}, requiring additional image information, and FSRCNN (Fast Super-Resolution Convolutional Neural Network) \cite{dong2016accelerating}. The highest visual quality is achieved with GAN-based models such as ESRGAN \cite{wang2018esrgan} and Real-ESRGAN \cite{wang2021realesrgan}. Besides the additional computational load required to evaluate the network, DL-based SR 
faces the problem of introducing artifacts, such as blurriness, unnatural textures, or `hallucinated' details that do not match the true image content if the networks are not properly trained. 

GaussianSR~\cite{hu2024gaussiansrhighfidelity2d} models images as continuous Gaussian fields to avoid pixel grid artifacts, and it can upscale images to any resolution without being limited to predefined scaling factors.
Compared to traditional interpolation methods, the method can preserve fine structures better and reduce aliasing and blocky artifacts found in pixel-based methods. 
On the downside, it requires end-to-end training of Gaussian kernels and neural networks, making it slower than traditional upscaling, and it currently is optimized for image-based super-resolution, lacking real-time support for interactive 3D rendering.


\paragraph{Super-resolution for novel view synthesis.} 
Neural Radiance Fields (NeRFs) \cite{mildenhall_nerf_2021} 
encode a 3D scene as a continuous function using a neural network and generate novel views by synthesizing the radiance at every point in 3D space.
Super-resolution layers can be integrated to upscale the lower-resolution NeRF output, enhancing resolution and finer details.
Such techniques are typically '3D-aware,' meaning they account for the 3D structure and depth information from the NeRF model when performing super-resolution.

NeRF-SR \cite{wang2022nerfsr} captures finer details and reduces aliasing artifacts by shooting multiple rays at each image pixel, combined with a refinement network that hallucinates details from related patches on a given high-resolution reference image. 
Using a discriminator network, GAN-based NeRF-SR models can add realistic textures to the upscaled output, making it appear more detailed and photorealistic.
Similarly, RefSR-NeRF \cite{huang2023refsrnerf} enhances NeRFs by integrating super-resolution capabilities guided by high-resolution reference images. 
It first constructs a NeRF model at a lower resolution and then performs super-resolution enhancement by utilizing a high-resolution reference image to reconstruct high-frequency details.

Mip-Splatting~\cite{Yu2024MipSplatting} improves 3DGS by addressing artifacts that arise from changing sampling rates, such as adjusting the focal length or camera distance.
Introducing a 3D smoothing filter constrains the Gaussian primitives' sizes based on the maximum sampling frequency of input views, thereby reducing high-frequency artifacts when zooming in. 
A box filter, which effectively reduces aliasing and dilation issues, is simulated by replacing the 2D dilation filter with a 2D Mip filter. 


Super-resolution 3DGS \cite{feng2024srgs} employs primitive densification to achieve a more detailed 3D representation. It also integrates a pre-trained 2D super-resolution model to capture fine details that guide Gaussian primitives for improved reconstruction. 
Recently, gradient-guided selective splitting has been introduced to refine pre-trained coarse Gaussian primitives into finer ones, inheriting properties from coarser scales via an encoded latent feature field \cite{xie2024supergs}.

While previous approaches operate within the 3D Gaussian space and require modifications to the training and rendering process, our method is designed to integrate seamlessly with all 3DGS optimization and rendering pipelines. 
To the best of our knowledge, we are the first to incorporate classical image upscaling directly into the optimization and rendering processes of 3DGS.


\section{Motivational Example}

Bicubic image upscaling interpolates between pixels using polynomial basis functions. 
In the following, we present a 1D example to motivate our upscaling approach by highlighting common limitations of polynomial interpolation and proposing ways to address them.

\begin{figure}
\centering
\begin{subfigure}{\linewidth}
    \includegraphics[width=\textwidth]{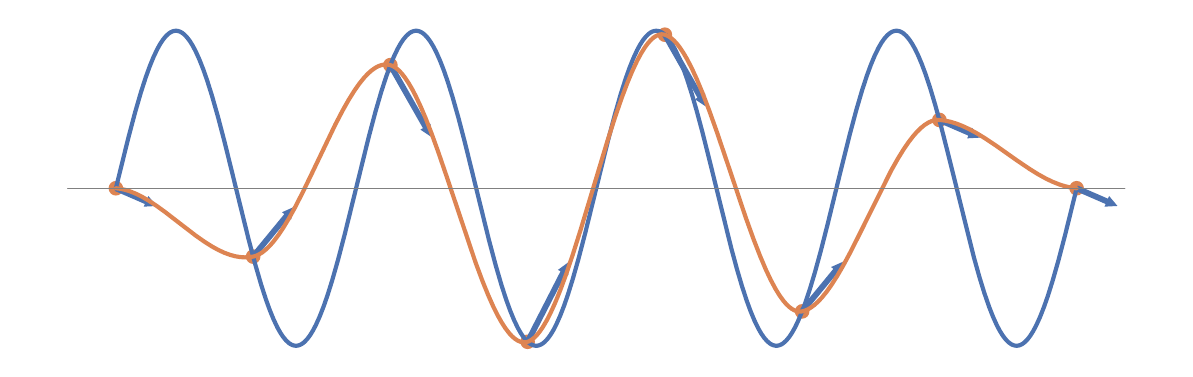}
    \caption{Finite difference approximation of gradients.}
    \label{fig:spline1d-finite}
\end{subfigure}
\begin{subfigure}{\linewidth}
    \includegraphics[width=\textwidth]{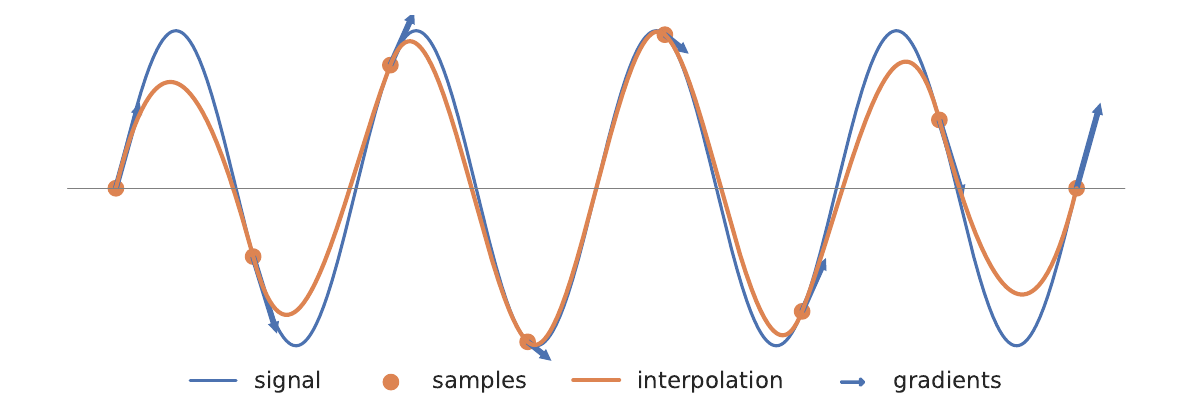}
    \caption{Analytical gradients.}
    \label{fig:spline1d-analytical}
\end{subfigure}
\caption{Reconstruction of a signal with cubic spline interpolation using gradients approximated via finite differences and analytical gradients.}
\label{fig:spline1d}
\vspace{-0.3cm}
\end{figure}

Consider a 1D signal sampled uniformly, i.e., multiplied by a 1D uniform comb function, analogous to how a rendering process samples objects along a scanline in the pixel raster (see~\cref{fig:spline1d}). 
Although originally continuous, the signal is represented by a discrete set of samples.
Interpolation methods like cubic, linear, or nearest-neighbor interpolation are commonly used to reconstruct the signal.
However, these traditional techniques often yield inaccurate approximations of the original signal, particularly when gradients are not well represented.

Cubic interpolation estimates the slope at each sample point using finite differences, which are then used to fit a smooth curve through the samples.
While finite differences provide an approximation of the gradient, they do not fully capture the continuous signal’s true rate of change.
Consequently, the reconstructed signal often diverges from the original, leading to undesirable artifacts (see~\cref{fig:spline1d-finite}).

To address this limitation, interpolation can be enhanced by sampling the signal itself and capturing its analytical gradients at each sample point. 
By storing gradient information alongside the sampled points, additional constraints can be applied during the interpolation process, enabling a more accurate spline fit.
Unlike traditional methods that rely solely on discrete data points and approximate gradients, our approach leverages the analytical gradients of Gaussian shape functions to more accurately replicate the signal’s shape, including subtle variations (see~\cref{fig:spline1d-analytical}).

\section{Differentiable 3D Gaussian Splatting}
\label{sec:prerequisites}

Differentiable 3DGS~\cite{kerbl_3d_2023} builds upon representing a 3D scene by a set of 3D Gaussians 
\begin{equation}\label{eq:3dgs}
    G(x) = \sigma e^{-\mu^T\Sigma^{-1}\mu}.
\end{equation}
A Gaussian is fully determined by its center $\mu\in\mathbb{R}^3$, a  covariance matrix $\Sigma \in\mathbb{R}^{3\times3}$ describing the Gaussian's orientation and shape, its opacity $\sigma \in [0,1]$, and a set of spherical harmonics (SH) coefficients to determine its view-dependent color. 
To render a scene, the 3D Gaussians are sorted in front-to-back order based on their mean $\mu$ and projected into the image plane using EWA Splatting~\cite{zwicker_ewa_2001}.
The 2D projection of a 3D Gaussian is again a Gaussian with covariance \begin{equation}
    \Sigma' = JW\Sigma W^TJ^T,
\end{equation}
where $W$ is the view transformation matrix and $J$ is the Jacobian of the affine approximation of the projective transformation~\cite{zwicker_ewa_2001}. This allows to evaluate the 2D color and opacity footprint of each projected Gaussian. A pixel's color $I(x,y)$ is then computed by blending all $N$ 2D Gaussians contributing to this pixel in sorted order: 
\begin{align}
    I(x,y) &= \sum_{i=1}^{N} T_i(x,y) \alpha_i(x,y) c_i \label{eq:3dgs-blend} \\
    \alpha_i(x,y) &= \sigma_i \exp(g_i(x,y)) \\
    T_i(x,y) &= \prod_{j=1}^{i-1}(1-\alpha_j(x,y)) 
\end{align}

Here, $c_i$ and $\sigma_i$ are the view-dependent color of a Gaussian and its opacity. 
$g_i(x,y)$ is derived from the 2D covariance matrix similar to \cref{eq:3dgs} in 3D.  


Initially proposed by Zwicker \etal~\cite{zwicker_ewa_2001} for modeling a 3D scalar field with 3D Gaussians, Kerbl \etal~\cite{kerbl_3d_2023} propose optimizing the Gaussians' parameters so that the rendering results match recorded images from the same view. 
Differentiable rendering regarding a color-based image loss is used for optimization, involving adaptive densification and removal of Gaussians to achieve a balance between reconstruction quality and the number of elements.

\section{Image Gradient Based Upscaling}

Our upscaling method fits a 
bicubic spline to a set of pixel values in the surrounding of the target pixel that should be interpolated. 
The spline assigns weights to each surrounding pixel based on its distance from the target pixel. Pixels closer to the target contribute more heavily, while those farther away contribute less. This weighting reduces the blocky appearance often seen in simpler interpolation methods, like nearest-neighbor or bilinear interpolation. In contrast to classical bicubic upscaling, we use the analytical gradients to determine the spline function.

\paragraph{Spline Interpolation}

\begin{figure}[t]
\centering
\includegraphics[width=0.4\linewidth]{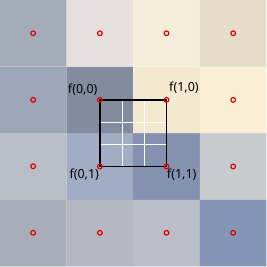}
\caption{Bicubic upscaling fits a bicubic polynomial $f(x,y)$ through a set of neighboring pixels. White quadrilaterals indicate target pixels in the upscaled image. For given pixel colors, the background colors show nearest neighbor interpolation.}
\label{fig:pixelmask}
\vspace{-0.3cm}
\end{figure}

Bicubic interpolation estimates a color at a target pixel by considering the colors of the 4x4 surrounding pixels, as shown in \cref{fig:pixelmask}. It is worth noting that the outer 12 pixels are used only for computing gradients in the color landscape.  

The coefficients $A\in \mathrm{R}^{4\times4}$ of the cubic interpolation polynomials are calculated by solving the system of linear equations $F = CAC^T$, with  
\begin{align}
F = \begin{bmatrix}
f(0,0)&f(0,1)&f_y(0,0)&f_y(0,1)\\
f(1,0)&f(1,1)&f_y(1,0)&f_y(1,1)\\
f_x(0,0)&f_x(0,1)&f_{xy}(0,0)&f_{xy}(0,1)\\
f_x(1,0)&f_x(1,1)&f_{xy}(1,0)&f_{xy}(1,1)\\
\end{bmatrix}.
\end{align}
$C \in \mathrm{R}^{4\times4}$ is a constant matrix and $f_x(x,y)$ are the derivatives of $f$ with respect to $x$, according to the setting in \cref{fig:pixelmask}.
In the supplemental material, we provide a thorough derivation of these quantities.

The polynomial can then be evaluated at an arbitrary point in the 2D pixel subdomain with corner points $f(0,0), f(1,0), f(0,1), f(1,1)$:
\begin{equation}
p(x,y) = \begin{bmatrix}1&x&x^2&x^3\end{bmatrix} A \begin{bmatrix}1&y&y^2&y^3\end{bmatrix}^T
\end{equation}
For each subdomain, a different matrix $A$ is computed and used for interpolation. 
For classical bicubic interpolation, the gradients of the pixel values are estimated with finite differences using the colors of neighboring pixels.


\paragraph{3DGS Image Gradients}

Compared to using finite differences for gradient approximation, a more accurate interpolation can be achieved by using the exact gradients of the smooth, continuous signal.
We propose to use the analytical gradients of the 2D Gaussians in image space instead of the finite difference approximations. In the following, we demonstrate how these gradients can be computed on the fly for an image generated with 3DGS.

When rendering a 3DGS model, each Gaussian is projected onto the camera image plane using EWA Splatting~\cite{zwicker_ewa_2001}.
Assuming $N$ Gaussians representing the scene, their 2D footprints are blended in sorted order into a final image $I(x,y)$ using alpha blending, as shown in \cref{eq:3dgs-blend}.

To compute the analytical gradients of the 2D color image, we differentiate the image $I(x,y)$ with respect to the 2D spatial coordinates $x$ and $y$:
\begin{align}
    \frac{\partial I(x,y)}{\partial x} &= \sum_{i=1}^N c_i (\frac{\partial T_i}{\partial x} \alpha_i + T_i \frac{\partial \alpha_i}{\partial x}) \\
    \frac{\partial I(x,y)}{\partial y} &= \sum_{i=1}^N c_i (\frac{\partial T_i}{\partial y} \alpha_i + T_i \frac{\partial \alpha_i}{\partial y}) \\
    \frac{\partial^2 I(x,y)}{\partial x\partial y} &=
    \begin{aligned}[t]
    \sum_{i=1}^N c_i (\frac{\partial^2 T_i}{\partial x\partial y} \alpha_i + \frac{\partial T_i}{\partial x} \frac{\partial \alpha_i}{\partial y} \\
    {} + \frac{\partial T_i}{\partial y} \frac{\partial \alpha_i}{\partial x} + T_i \frac{\partial^2 \alpha_i}{\partial x\partial y})
    \end{aligned}
\end{align}

To compute the derivative of the transmittance $T_i$ with respect to the screen position $x$, $T_i$ is first reformulated by means of the $\alpha$-blending function:
\begin{align}
    A_i &= A_{i-1} + \alpha_i(1-A_{i-1}) \\
    T_i &= 1-A_{i-1} 
\end{align}

This formulation allows the derivative to be computed iteratively as part of the Gaussian blending operation:
\begin{align}
    \frac{\partial T_i}{\partial x} &= - \frac{\partial A_{i-1}}{\partial x}\\
    \frac{\partial A_i}{\partial x} &= \frac{\partial A_{i-1}}{\partial x}  (1-\alpha_i) + (1-A_{i-1}) \frac{\partial \alpha_i}{\partial x}
\end{align}

The calculation of $\frac{\partial T_i}{\partial y}$ and $\frac{\partial^2 T_i}{\partial x\partial y}$ is analogous. The derivation of all partial derivatives is provided in the supplemental material.

As shown, the gradients in image space can be obtained by blending weighted per-Gaussian gradients, which can be computed analytically due to the differentiability of the Gaussian kernel functions. In particular, the computations can be included in the 3DGS rendering pipeline with only minor overhead (cf.\ rasterization timings in \cref{{tab:timing-forward}}). Additional details regarding the calculation of analytical gradients from a given 3DGS model are provided in the supplemental material.

\begin{figure}[]
    \centering
    \includegraphics[width=\linewidth]{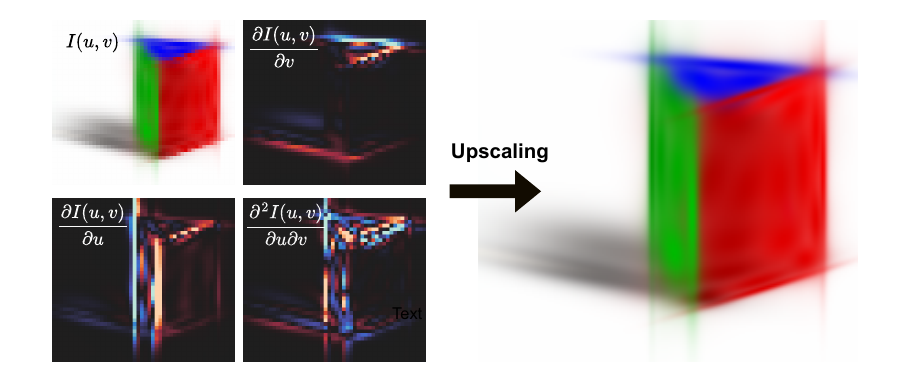}
    \caption{Our differentiable 3DGS rasterizer produces a low resolution image and its analytical gradients in image space. Gradients are used to upscale the image with spline interpolation to the target resolution.}
    \label{fig:pipeline}
    \vspace{-0.4cm}
\end{figure}

In the 3DGS forward rendering pass, the analytical gradients are available at every pixel of the used pixel raster. These gradients are then used as outlined to perform spline-based image interpolation. The fully differentiable 3DGS rendering process , including image upscaling, is illustrated in \cref{fig:pipeline}.

We perform a first quality assessment of spline-based interpolation using 2D images commonly used for image interpolation benchmarking. To generate a lower-resolution input, we use a 2D Gaussian model trained to represent a 2D color image and render it at a reduced resolution. The rendered image is then upscaled using bicubic spline-based interpolation. 

\begin{figure}[h]
    \includegraphics[width=\linewidth]{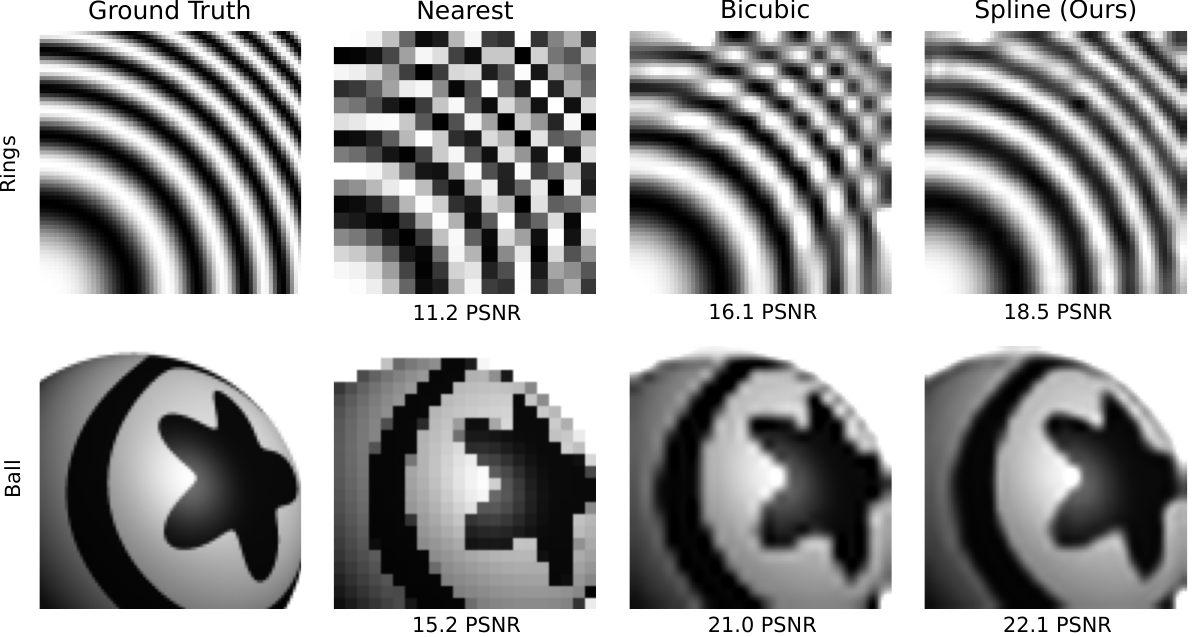}
    \caption{Interpolation benchmarking. A 2D Gaussian Splatting model is fitted to the ground truth image, and upscaling methods are applied to the image rendered at $1/4$ the resolution of the original. Images proposed by \cite{getreuter_2011} for interpolation benchmarking. }
    \label{fig:2dgs-upscale}
    \vspace{-0.4cm}
\end{figure}

The results are shown in~\cref{fig:2dgs-upscale}. While bicubic interpolation introduces artifacts such as ringing and staircase effects, our method effectively mitigates these distortions, producing smoother and more visually accurate results.


\paragraph{Differentiable Image Gradients}

While front-to-back blending is performed in the forward pass, 3DGS performs back-to-front blending in the 
backward pass to efficiently compute the gradients. 
Using back-to-front blending, the image $I(x,y)$ is computed as
\begin{align}
    I(x,y) &= B_1 \\
    B_{i} &= (1-\alpha_i)B_{i+1} + c_i \alpha_i
\end{align}

As an example for other parameters, we demonstrate the calculation of the partial derivative of the image gradient with respect to opacity $\sigma$. The calculations for other parameters are analogous and can be found in the supplemental material. Using the chain rule, the gradient with respect to $\sigma_k$ of the $k$-th Gaussian is formulated as follows:
\begin{align}
    \frac{\partial I(x,y)}{\partial \sigma_k} &= \frac{\partial B_1}{\partial \sigma_k} = \frac{\partial B_1}{\partial \alpha_k} \frac{\partial \alpha_k}{\partial \sigma_k} \\
    \frac{\partial^2 B_i}{\partial x \partial \sigma_k } &= \frac{\partial^2 B_i}{\partial \alpha_k \partial x} \frac{\partial \alpha_k}{\partial \sigma_k} + \frac{\partial B_i}{\partial \alpha_k} \frac{\partial^2 \alpha_k}{\partial \sigma_k \partial x}
\end{align}

From this formulation, compact solutions for the partial derivatives can be derived, which can be calculated within the iterative back-to-front blending loop:
\begin{align}
    \frac{\partial I(x,y)}{\partial \alpha_k} &=\frac{\partial B_1}{\partial \alpha_k} = (1-A_{k-1}) (c_k - B_{k+1}) \\
    \frac{\partial^2 I(x,y)}{\partial \alpha_k \partial x} &= -\frac{\partial A_{k-1}}{\partial x} (c_k-B_{k+1}) - \frac{\partial B_{k+1}}{\partial x}(1-A_{k-1}) 
\end{align}

Note that $A_{k-1} = 1 - T_i$ is the alpha blending term calculated with front-to-back blending.
We can compute this term for each iteration in the backward pass using the Inversion-Trick~\cite{weiss_differentiable_2022}. 
We provide a detailed derivation of the gradient formulas in the supplemental material. 

\begin{table*}[ht!]
\resizebox{\linewidth}{!}{
\begin{tabular}{rr|SSSSS|SSSSS|SSSSS}
\toprule
{} & {} & \multicolumn{5}{c|}{MipNeRF360 (5187x3361)} & \multicolumn{5}{c|}{Tanks and Temples (979x546)} & \multicolumn{5}{c}{Deep Blending (1264x832)} \\
{Scale} & {Method} & {{SSIM $\uparrow$}} & {{PSNR $\uparrow$}} & {{LPIPS $\downarrow$}} & {{FPS $\uparrow$}} & {{Train $\downarrow$}} & {{SSIM $\uparrow$}} & {{PSNR $\uparrow$}} & {{LPIPS $\downarrow$}} & {{FPS $\uparrow$}} & {{Train $\downarrow$}} & {{SSIM $\uparrow$}} & {{PSNR $\uparrow$}} & {{LPIPS $\downarrow$}} & {{FPS $\uparrow$}} & {{Train $\downarrow$}} \\
\midrule
 \midrule\multirow[c]{5}{*}{2} & 3DGS & 0.809 & 26.97 &  0.308 & 15 & {\cellcolor{best}} 105 & 0.799 & 22.09 & 0.220 & 156 & {\cellcolor{best}} 17 & 0.901 & 29.64 & 0.247 & 209 & {\cellcolor{best}} 17 \\
 & Mip~\cite{Yu2024MipSplatting} & {\cellcolor{best}} 0.815 & {\cellcolor{best}} 27.23 & {\cellcolor{best}} 0.302 & 10 & 143 & 0.811 & 22.62 & {\cellcolor{best}} 0.190 & 126 & 20 & {\cellcolor{best}} 0.908 & {\cellcolor{best}} 29.95 & {\cellcolor{best}} 0.239 & 118 & 22 \\
 & Bicubic &  0.813 & 26.94 & 0.310 & {\cellcolor{best}} 43 & 123 & 0.799 & 22.19 & 0.238 & 213 & 18 & 0.902 & 29.67 & 0.249 & 429 & 18 \\
 & Spline (Ours) & 0.813 & 27.00 & 0.310 & 41 & 139 & {\cellcolor{best}} 0.826 & {\cellcolor{best}} 22.64 & 0.206 & {\cellcolor{best}} 235 & 18 & 0.905 & 29.69 & 0.245 & {\cellcolor{best}} 430 & 18 \\
  & Spline (3DGS) &  0.809 & 26.94 & 0.309 & 35 & 105 & 0.783 & 21.93 & 0.239 & 186 & 17 & 0.905 & 29.76 & 0.246 & 313 & 17 \\
 \midrule\multirow[c]{5}{*}{4} & 3DGS & 0.783 & 26.35 & 0.347 & 15 & {\cellcolor{best}} 59 & 0.677 & 20.31 & 0.341 & 142 & {\cellcolor{best}} 19 & 0.877 & 28.89 &  0.286 & 209 & {\cellcolor{best}} 14 \\
  & Mip~\cite{Yu2024MipSplatting} & 0.806 &{\cellcolor{best}} 27.14 & {\cellcolor{best}} 0.317 & 10 & 73 & 0.705 & 20.85 & {\cellcolor{best}} 0.293 & 116 & 20 & {\cellcolor{best}}0.896 & {\cellcolor{best}}29.56 &{\cellcolor{best}} 0.265 & 112 & 16 \\
 & Bicubic & 0.799 & 26.60 & 0.338 & {\cellcolor{best}} 90 & 83 & 0.677 & 20.28 & 0.392 & 124 & 20 & 0.875 & 28.70 & 0.312 & 399 & 15 \\
 & Spline (Ours) & {\cellcolor{best}} 0.809 & 26.85 & 0.322 & 90 & 91 & {\cellcolor{best}} 0.744 & {\cellcolor{best}} 21.30 & 0.323 & {\cellcolor{best}} 159 & 20 &  0.890 & 29.09 & 0.286 & {\cellcolor{best}} 423 & 16 \\
& Spline (3DGS) & 0.776 & 25.71 & 0.342 & 69 & 59 & 0.654 & 20.03 & 0.379 & 122 & 19 & 0.886 & 29.14 & 0.291 & 312 & 14 \\
\bottomrule
\end{tabular}
}

\caption{Comparison of model quality and rendering speed w and w/o upscaling during training (training times in minutes) and rendering. For MipNerf360~\cite{barron_mip-nerf_2022}, 5 million Gaussians are used, and 1 million Gaussians otherwise. \textit{3DGS} and \textit{Mip} train with reduced image resolution and render to full resolution. \textit{Bicubic} upscales low-resolution renderings from 3DGS. 
\textit{Spline (Ours)} includes upscaling during training, renders to low-resolution and upscales. \textit{Spline (3DGS)} trains with reduced image resolution, renders to low-resolution and upscales.}
\label{tab:train-results}
\vspace{-0.2cm}
\end{table*}

\paragraph{Scene Reconstruction}
Since both the image gradient calculation and spline-based interpolation are fully differentiable with respect to the scene parameters, image upscaling using analytical gradients can be seamlessly integrated into the optimization pipeline for reconstructing the 3DGS model.
For bicubic interpolation, the pipeline remains identical, except that gradients are computed using finite differences during the interpolation stage.


During optimization, the 3DGS renderer runs at a lower resolution, with the rendered images subsequently upscaled to match the ground-truth image resolution. Thus, the rasterizer needs to generate significantly fewer fragments, vastly reducing rendering times in the backpropagation pass. Image gradients are then backpropagated to the Gaussians by chaining the image derivatives with respect to spline upscaling and analytical gradient computation.



\section{Experiments}

\subsection{Datasets}

We evaluate our method on the \textbf{Mip-Nerf360~\cite{barron_mip-nerf_2022}} indoor and outdoor scenes, and two scenes from the \textbf{Tanks\&Temples \cite{knapitsch_tanks_2017}} and \textbf{Deep Blending~\cite{hedman_deep_2018}} dataset.
For all experiments, we used the full-resolution images provided by the datasets.

\subsection{Implementation Details}

Our implementation is based on the 3DGS code by Kerbl~\etal~\cite{kerbl_3d_2023}.
Rendering and training speed heavily depends on the number of Gaussians.
To ensure a fair comparison between methods, we use the densification method proposed by Mallick~\etal~\cite{taming3dgs}, where the number of Gaussians grows according to a predefined function during training.

For the gradient-aware upscaling, we manually derived the forward and backward pass and implemented it in CUDA.
The spline image upscaler is implemented in SLANG.D~\cite{bangaru2023slangd} using its CUDA backend. 
For bicubic interpolation, the differentiable bicubic interpolation provided by PyTorch~\cite{paszke2017automatic} is used.
Further details are provided in the supplemental material.
All experiments were conducted using NVIDIA L4 GPUs unless otherwise specified.

\subsection{Results}

Our evaluations include training 3DGS at various resolutions (half, quarter and eighth) to assess how rendering at different scales impacts quality and performance.
We apply bicubic upscaling (2x, 4x, 8x) as a baseline comparison and introduce our spline-based upscaling method (also at 2x, 4x, 8x) to improve image quality.
We refrain from using Lanczos upscaling in the following evaluations, as it introduces a significant computational overhead that does not translate into improved image quality. 
We demonstrate this property in the supplemental material.

We compare the visual results and timings for all training configurations to comprehensively evaluate each approach. 
Additionally, we measure the inference render time of each method to assess their performance in real-time applications. 
All images and metrics presented in the paper are derived from the respective test sets of the datasets.



\paragraph{Scene Reconstruction}

Spline-based upscaling consistently outperforms bicubic interpolation in visual quality and produces better outcomes when rendering a 3DGS model at lower resolutions and then upscaling to the target resolution. 
\cref{tab:train-results} shows PSNR, Structural Similarity Index Measure (SSIM), and LPIPS metrics for the different upscaling methods across datasets.
Our method performs particularly well on the LPIPS metric, which better aligns with human perception. 

In the supplemental material, we demonstrate that spline-based upscaling achieves a good trade-off between quality and training time, including metrics for individual scenes.
It can be observed that bicubic upscaling is minimally faster than spline-based upscaling during training but yields worse results in all experiments.

\begin{figure*}
    \centering
    \includegraphics[width=\textwidth]{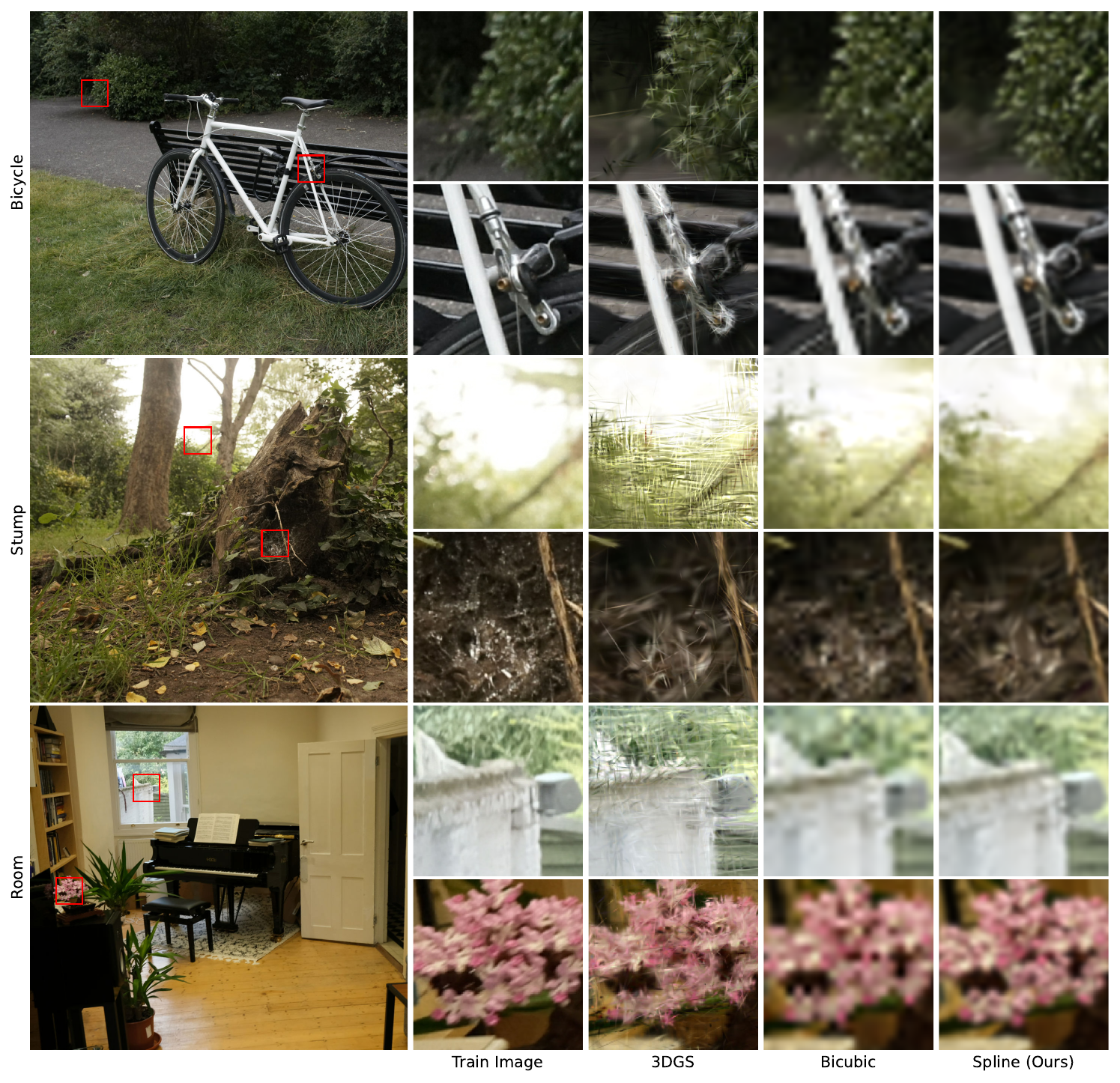}
    \caption{
    Image quality comparison. 3DGS model trained on $\frac{1}{8}$ the training image resolution.
    3DGS rendered on full resolution (3DGS), and $8\times$ upscaling and rendering during training with bicubic (Bicubic) and our spline-based (Spline) interpolation.
    }
    \label{fig:comparison-crop}
\end{figure*}

The visual quality improvements of spline-based upscaling are also supported by the results  in \cref{fig:comparison-crop}. 
Our method produces sharper images with fewer aliasing artifacts than bicubic interpolation.
Compared to full-resolution 3DGS rendering, spline-based upscaling—when trained on low-resolution input—produces a perceptually improved output, avoiding needle artifacts that become especially noticeable as popping in animations. Notably, this is achieved at significantly improved rendering times.

The improvements in training speed are not as significant as expected.
The primary computational bottleneck is the evaluation of the SSIM loss function on the full-resolution image and the backward pass of the upscaling function. 
Since these evaluations require a significant portion of the computational workload, the performance gains from reduced rasterization load cannot fully unfold their potential.
Detailed timings can be found in the supplementary material.

\paragraph{Rendering}
 \cref{tab:timing-forward} shows that spline-based upscaling is only marginally slower than bicubic upscaling, yet achieves significantly higher quality, as demonstrated. Compared to 3DGS rendering at full resolution, 3x-4x performance increases are achieved, regardless on which resolution the 3DGS model has been trained.  
The timings indicate the suitability of Spline-based upscaling for real-time rendering applications, especially on low-end hardware where full-resolution rendering prohibits interactive frame rates.


\begin{table}[h]
\centering
\begin{tabular}{l|ccc}
Method & Render & Upscale & Speedup \\ \midrule
3DGS & 72.4 ms & - & - \\
Bicubic & 15.4 ms & 1.8 ms & $\times4.2$ \greenup \\
DLSS & 15.4 ms & 8.0 ms & $\times3.1$ \greenup \\
NinaSR-B1 & 15.4 ms & 275.3 ms & $\times0.2$ \reddown \\
Spline (Ours) & 15.6 ms & 1.8 ms & $\times4.2$ \greenup \\
\end{tabular}%
\caption{Performance comparison of rendering and upscaling (milliseconds per frame) for `garden' at  target resolution $4096 \times 2304$ using 5 Mio. Gaussians. 3DGS renders to the target resolution; all others perform $3\times$ upscaling of the low-resolution rendering.}
\label{tab:timing-forward}
\vspace{-0.3cm}
\end{table}

\begin{figure*}[h!]
    \centering
    \begin{subfigure}{0.19\linewidth}
        \begin{tikzpicture}
            \draw (0, 0) node[inner sep=0] {\includegraphics[width=\linewidth]{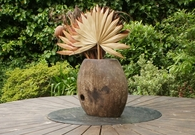}};
        \end{tikzpicture}
        \caption{Image from test set}
        \label{fig:comp-fullres}
    \end{subfigure}
    \hfill
    \begin{subfigure}{0.19\linewidth}
        \begin{tikzpicture}
            \draw (0, 0) node[inner sep=0] {\includegraphics[width=\linewidth]{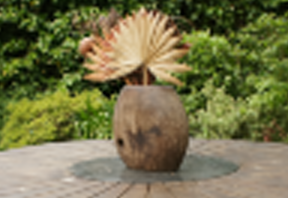}};
            \node[fill=black, text=white,  anchor=north east, inner sep=5pt, font=\footnotesize, inner sep=4pt] at (0.5\linewidth,1.14) {$1.8$ms};
        \end{tikzpicture}
        \caption{Bicubic}
        \label{fig:comp-bicubic}
    \end{subfigure}
    \hfill
    \begin{subfigure}{0.19\linewidth}
        \begin{tikzpicture}
            \draw (0, 0) node[inner sep=0] {\includegraphics[width=\linewidth]{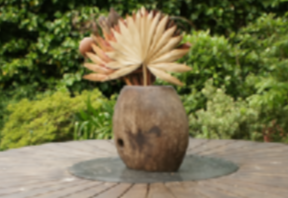}};
            \node[fill=black, text=white,  anchor=north east, inner sep=5pt, font=\footnotesize, inner sep=4pt] at (0.5\linewidth,1.14) {$1.8$ms};
        \end{tikzpicture}
        \caption{Spline (ours)}
    \end{subfigure}
    \hfill
    \begin{subfigure}{0.19\linewidth}
        \begin{tikzpicture}
            \draw (0, 0) node[inner sep=0] {\includegraphics[width=\linewidth]{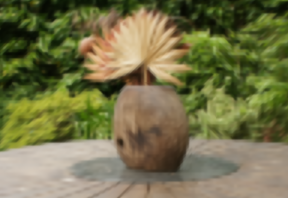}};
            \node[fill=black, text=white,  anchor=north east, inner sep=5pt, font=\footnotesize, inner sep=4pt] at (0.5\linewidth,1.14) {$8.0$ms};
        \end{tikzpicture}
        \caption{DLSS~\cite{nvidia_dlss}}
    \end{subfigure}
    \hfill
    \begin{subfigure}{0.19\linewidth}
        \begin{tikzpicture}
            \draw (0, 0) node[inner sep=0] {\includegraphics[width=\linewidth]{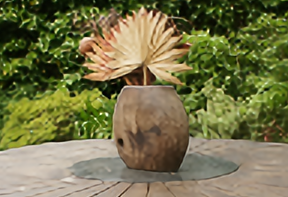}};
            \node[fill=black, text=white,  anchor=north east, inner sep=5pt, font=\footnotesize, inner sep=4pt] at (0.5\linewidth,1.14) {$275.3$ms};
        \end{tikzpicture}
        \caption{NinaSR-B1 \cite{ninasr}}
    \end{subfigure} 
    \caption{Comparison of $3\times$ upscaling. Upscale time in milliseconds for 4K resolution.
    }
    \label{fig:figures}
\end{figure*}

\paragraph{Deep Learning Based Upscaling}
One of the most competitive DL-based upscalers, NinaSR-B1~\cite{ninasr}, produces sharper  details than ours (cf.~\cref{fig:figures}), yet at significantly lower performance. 
Furthermore, spline-based upscaling does not require exhaustive training and maintains interpretability by avoiding the 'hallucinated' details often introduced by deep learning methods.
In the supplemental material, we provide additional results for DL-based upscaling in the training process using NinaSR-B1.

NVIDIA's DLSS~\cite{nvidia_dlss} even shows a loss in image quality compared to spline-based upscaling, with a tendency to overblur the image.
We attribute this to the absence of an accurate depth map, which would be required to fully exploit DLSS's potential.
Our approach upscales at competitive speed without requiring this additional information.

Spline-based upscaling also provides temporal stability, ensuring frame-to-frame consistency, which is critical in interactive scenarios (see supplemental material). In contrast, DL-based approaches struggle to achieve temporal consistency and require additional data, such as geometry- and image-based motion vectors—information that cannot be accurately computed for 3DGS models.

\subsection{Limitations}

Our approach is designed explicitly for 3DGS, where analytical gradients can be computed efficiently.
This advantage is unlikely to generalize to other novel view synthesis methods, such as NeRFs, which do not share the same efficiency in gradient computation and may not benefit from spline-based upscaling.

Our approach reduces artifacts of bicubic interpolation and generates smoother and more visually pleasing images. 
However, we cannot reconstruct sharp details due to the smooth nature of the splines used for interpolation. 
The use of higher-order splines and alternative basis functions may help mitigate this limitation. 


\section{Conclusion}
We introduce a new method for upscaling 3DGS images that employ analytical gradients of Gaussian kernel functions to achieve a better fit of a cubic interpolation polynomial.
Our extensive experiments show that our approach delivers high-resolution images in real-time while improving over classical image upscalers and existing real-time DL upscalers such as DLSS~\cite{nvidia_dlss}.

By training on higher-resolution images without requiring full-resolution rendering, we reduce the rendering time, offering a better quality-time tradeoff when compared to 3DGS~\cite{kerbl_3d_2023}.

In future work, we will explore spline-based upscaling of images generated with other kernels functions~\cite{chen2024linear} and, in particular, ray-tracing~\cite{3dgrt2024} to reduce the number of costly rays.
Furthermore, we intend to combine spline-based upscaling with hardware rasterization to further increase rendering performance.



\clearpage

{
    \small
    \bibliographystyle{ieeenat_fullname}
    \bibliography{main}
}
\input{sec/X_suppl}

\end{document}

%% file: authors.tex
\def\tumonly{1}

\if\tumonly1

\author{Simon Niedermayr\\
{\tt\small simon.niedermayr@tum.de}
\and
Christoph Neuhauser\\
{\tt\small christoph.neuhauser@tum.de}
\and
Rüdiger Westermann\\
{\tt\small westermann@tum.de} \\
\and
{Technical University of Munich} 
}

\else

\author{Simon Niedermayr\\
Technical University of Munich\\
{\tt\small simon.niedermayr@tum.de}
\and
Christoph Neuhauser\\
Intel Corporation\\
{\tt\small christoph.neuhauser@intel.com}
\and
Rüdiger Westermann\\
Technical University of Munich\\
{\tt\small westermann@tum.de}
}

\fi

%% file: sec/X_suppl.tex
\maketitlesupplementary

\renewcommand\thesubsection{\Alph{subsection}}

\subsection{Qualitative Results for Different Upsampling Rates}

Our method enables image upscaling with arbitrary factors, including fractional values, offering flexibility beyond fixed integer magnifications.
While our paper primarily showcases visual results for higher upscaling factors (4x,8x), our approach also demonstrates superior performance at lower magnifications, such as 2× or 3×, when compared to traditional interpolation techniques like bicubic or Lanczos (see~\cref{fig:comparison-crop}).
Specifically, our method reduces artifacts and preserves finer details more effectively.

However, these improvements are most apparent when the upscaled image is displayed at its native resolution, where the screen’s pixel grid aligns with the image pixels.
When viewed in formats like PDFs, where rendering software may resample, process, or scale images dynamically, these advantages can become less noticeable due to unintended alterations introduced by the viewer.
In~\cref{fig:comparison-crop}, we show qualitative results for 3x upscaling.

\subsection{Performance Analysis}

In this section, we provide a more detailed performance analysis of our approach. 
\cref{tab:render_times_all} presents rendering times for different upscaling methods and upscaling factors, highlighting the computational cost of each configuration.

\begin{figure}[h]
    \centering
    \includegraphics[width=0.95\linewidth]{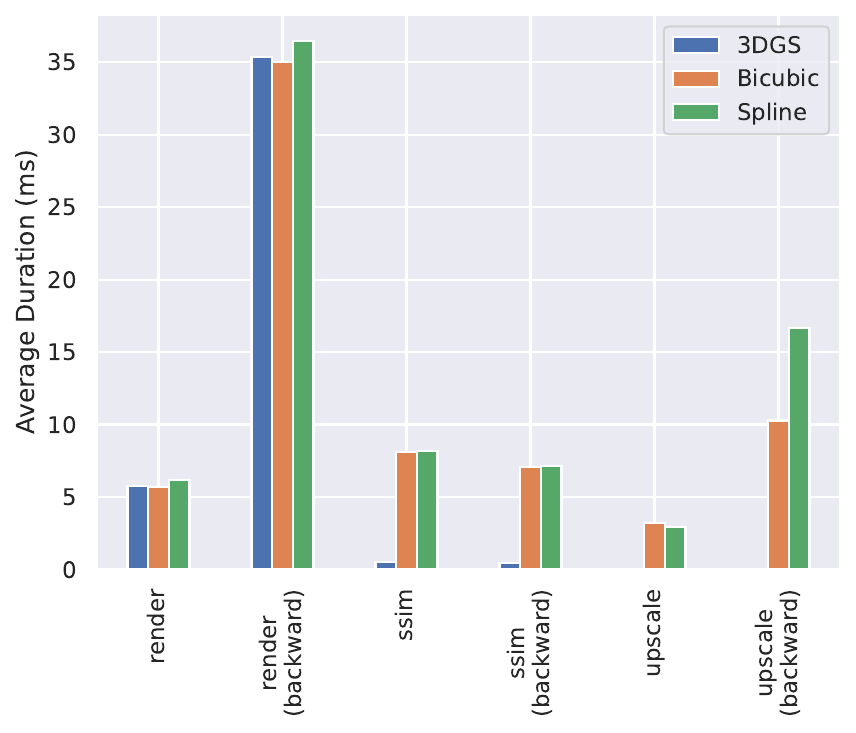}
    \caption{Averge time of different operations in the training pipeline. The MipNeRF360 garden scene with 5 Million Gaussians and the full resolution was used to obtain measurements. 3DGS was trained at $\frac{1}{4}$ resolution while Spline and Bicubic were trained with 4x upscaling. }
    \label{fig:training-timings}
\end{figure}

Additionally, we analyze the performance during training, focusing on the timing of key operations.
As shown in~\cref{fig:training-timings} we measure the time required for the forward and backward pass of the most computationally expensive functions: rendering, SSIM computation, and upscaling. 
Our results indicate that the gradient computation has a negligible impact on the performance of the render function.

However, computing SSIM at a higher resolution (4×) is significantly more expensive.
Furthermore, the backward pass of our upscaling function is relatively slow and constitutes a major bottleneck. 
This is particularly evident in our spline upscaler, which relies on a naive implementation that extensively uses atomic operations in the backward pass.
Using subgroup operations and more efficient synchronization methods could significantly improve performance.

\subsection{Comparison to Mip Splatting}

We also evaluated Mip-Splatting~\cite{Yu2024MipSplatting} in comparison to our approach. Mip Splatting introduces a 3D filter that accounts for the scene’s sampling rate (training image resolution).
By dilating the Gaussians during the reconstruction based on the sampling rate, Mip Splatting effectively reduces straw effects that can otherwise be seen in 3DGS reconstruction.

However, this filtering process enlarges the Gaussians and decreases their opacity.
This affects the rendering process in two key ways. 
First, the increased Gaussian size results in more Gaussians contributing to each pixel, thereby increasing rendering and training time.
Second, the reduced opacity diminishes the effectiveness of early termination in the rendering pipeline. 
Since early termination typically halts processing more Gaussians when a pixel’s alpha value nears one, lower opacity Gaussians require more blending steps before reaching full coverage, thereby increasing the overall rendering and training time (see~\cref{tab:mip-metrics}).

While Mip Splatting  improves image consistency at high resolutions, its computational overhead makes it less suitable for real-time applications on low-end devices. 
Nevertheless, our upscaling method can be applied to scenes reconstructed with Mip Splatting, thereby speeding up the rendering process.

\begin{table}[h]
\centering
\begin{adjustbox}{width=\linewidth}
\begin{tabular}{llSSSSS}
\toprule
{Scale} & {Method} & {{SSIM $\uparrow$}} & {{PSNR $\uparrow$}} & {{LPIPS $\downarrow$}} & {{FPS $\downarrow$}} & {{Train $\downarrow$}} \\
\midrule
 \midrule\multirow[c]{3}{*}{2} & 3DGS & 0.809 & 26.97 & 0.308 & 15 & 105 \\
 & Mip & 0.815 & 27.23 & 0.302 & 10 & 143 \\
 & Spline (Ours) & 0.813 & 27.00 & 0.31 & 41 & 139 \\
 \midrule\multirow[c]{3}{*}{4} & 3DGS & 0.783 & 26.35 & 0.347 & 15 & 59 \\
 & Mip & 0.806 & 27.14 & 0.317 & 10 & 73 \\
 & Spline (Ours) & 0.809 & 26.85 & 0.322 & 90 & 91 \\
 \midrule\multirow[c]{3}{*}{8} & 3DGS & 0.681 & 24.67 & 0.446 & 14 & 48 \\
 & Mip & 0.775 & 26.37 & 0.371 & 8 & 55 \\
 & Spline (Ours) & 0.776 & 25.95 & 0.385 & 122 & 80 \\
\bottomrule
\end{tabular}

\end{adjustbox}
\caption{Metrics for Mip-Splatting~\cite{Yu2024MipSplatting} compared to baseline 3DGS~\cite{kerbl_3d_2023} and our Spline Upscaler. Average for all scenes in MipNerf360~\cite{barron_mip-nerf_2022} is reported. Metrics are evaluated at full resolution (5187x3361 pixels). 3DGS and Mip were trained at lower resolution (scale) while Spline (ours) uses upscaling to match ground truth image resolution during training.}
\label{tab:mip-metrics}
\end{table}

\subsection{Lanczos Image Interpolation}

We compare Lanczos and bicubic interpolation for image upscaling. Lanczos, though theoretically superior in edge preservation~\cite{getreuter_2011}, showed no major improvements over bicubic in our experiments. Additionally, it is more computationally expensive to compute.

\begin{figure*}[h]
    \centering
    \begin{subfigure}{0.30\linewidth}
        \includegraphics[width=\linewidth]{img/cutout_upscale/bicubic.png}
        \caption{Bicubic}
    \end{subfigure}
    \hfill
    \begin{subfigure}{0.30\linewidth}
        \includegraphics[width=\linewidth]{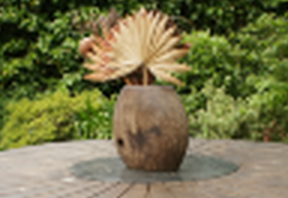}
        \caption{Lanczos}
    \end{subfigure}
    \hfill
    \begin{subfigure}{0.30\linewidth}
        \includegraphics[width=\linewidth]{img/cutout_upscale/spline.png}
        \caption{Spline (Ours)}
    \end{subfigure}
    \caption{Comparison of $4\times$ upscaling for different upscaling methods. Bicubic and Lanczos observe staircase and rining artifacts which are not present in Spline upscaling.}
    \label{fig:comp_lanczos}
\end{figure*}

Given the lack of significant visual benefits, bicubic remains the preferable choice as the baseline in our experiments. Visual comparisons can be found in Figure~\cref{fig:comp_lanczos}.

\begin{figure*}[h!]
    \centering
    \begin{subfigure}{0.49\linewidth}
        \setlength{\lineskip}{0pt}
        \begin{tikzpicture}
        \node[anchor=south west,inner sep=0] (image) at (0,0) {\includegraphics[width=1\linewidth]{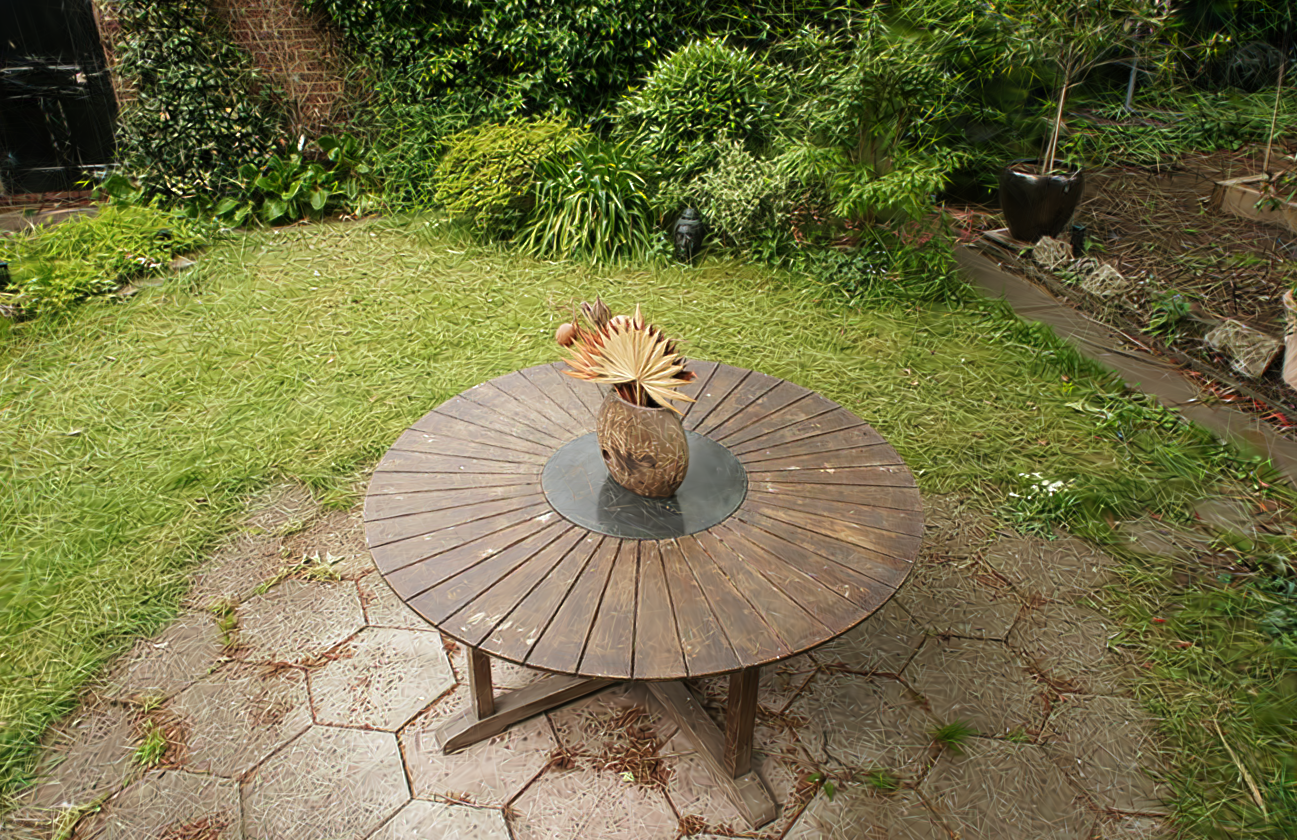}};
        \begin{scope}[x={(image.south east)},y={(image.north west)}]
            \draw[red,thick] (0.4027777777777778,0.6591179976162098) rectangle (0.5725308641975309,0.8617401668653158);
            \draw[red,thick] (0.4027777777777778,0.45172824791418353) rectangle (0.5725308641975309,0.6543504171632897);
        \end{scope}
        \end{tikzpicture}\vspace{0.1cm}
        \includegraphics[width=0.492\linewidth]{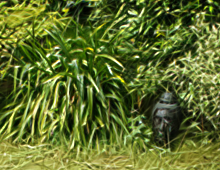}\hfill
        \includegraphics[width=0.492\linewidth]{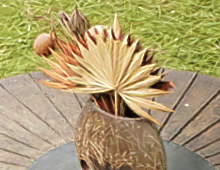}
        \caption{NinaSR-B1 $3\times$, PSNR = 21.61, 128min (train)}
    \end{subfigure}\hfill
    \begin{subfigure}{0.49\linewidth}
        \setlength{\lineskip}{0pt}
        \begin{tikzpicture}
        \node[anchor=south west,inner sep=0] at (0,0) {\includegraphics[width=1\linewidth]{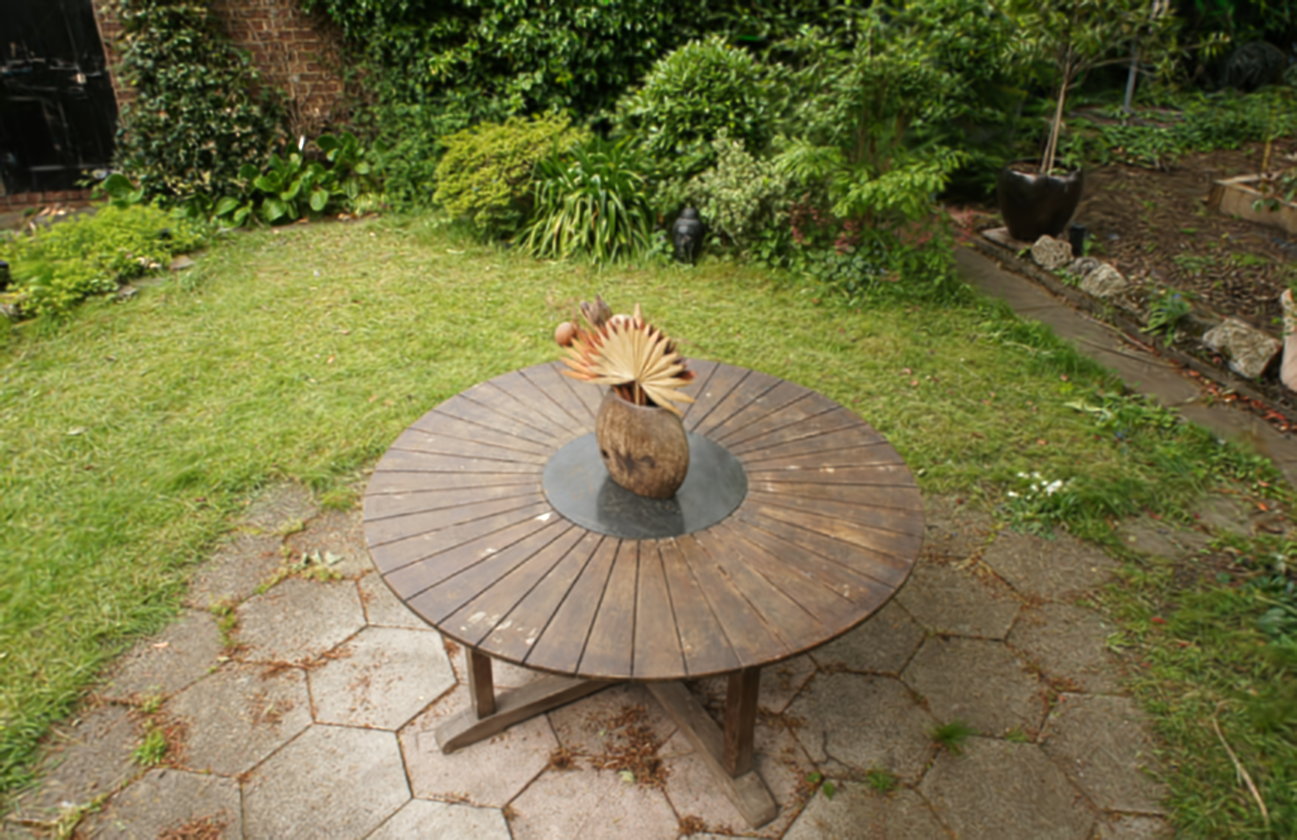}};
        \begin{scope}[x={(image.south east)},y={(image.north west)}]
            \draw[red,thick] (0.4027777777777778,0.6591179976162098) rectangle (0.5725308641975309,0.8617401668653158);
            \draw[red,thick] (0.4027777777777778,0.45172824791418353) rectangle (0.5725308641975309,0.6543504171632897);
        \end{scope}
        \end{tikzpicture}\vspace{0.1cm}
        \includegraphics[width=0.492\linewidth]{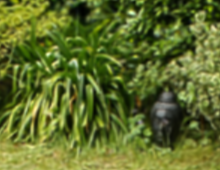}\hfill
        \includegraphics[width=0.492\linewidth]{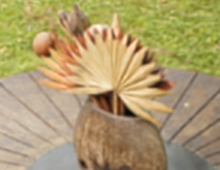}
        \caption{Spline $3\times$ (ours), PSNR = 25.30, 51min (train)}
    \end{subfigure}

    \caption{Comparison of training results for the Garden scene from Mip-NeRF~\cite{barron_mip-nerf_2022}.  Training image resolutions of $1296\times840$ were used.}
    \label{fig:ninasr-train}
\end{figure*}

\subsection{Detailed Metrics}

\begin{figure*}[ht]
    \centering
    \includegraphics[width=0.99\textwidth]{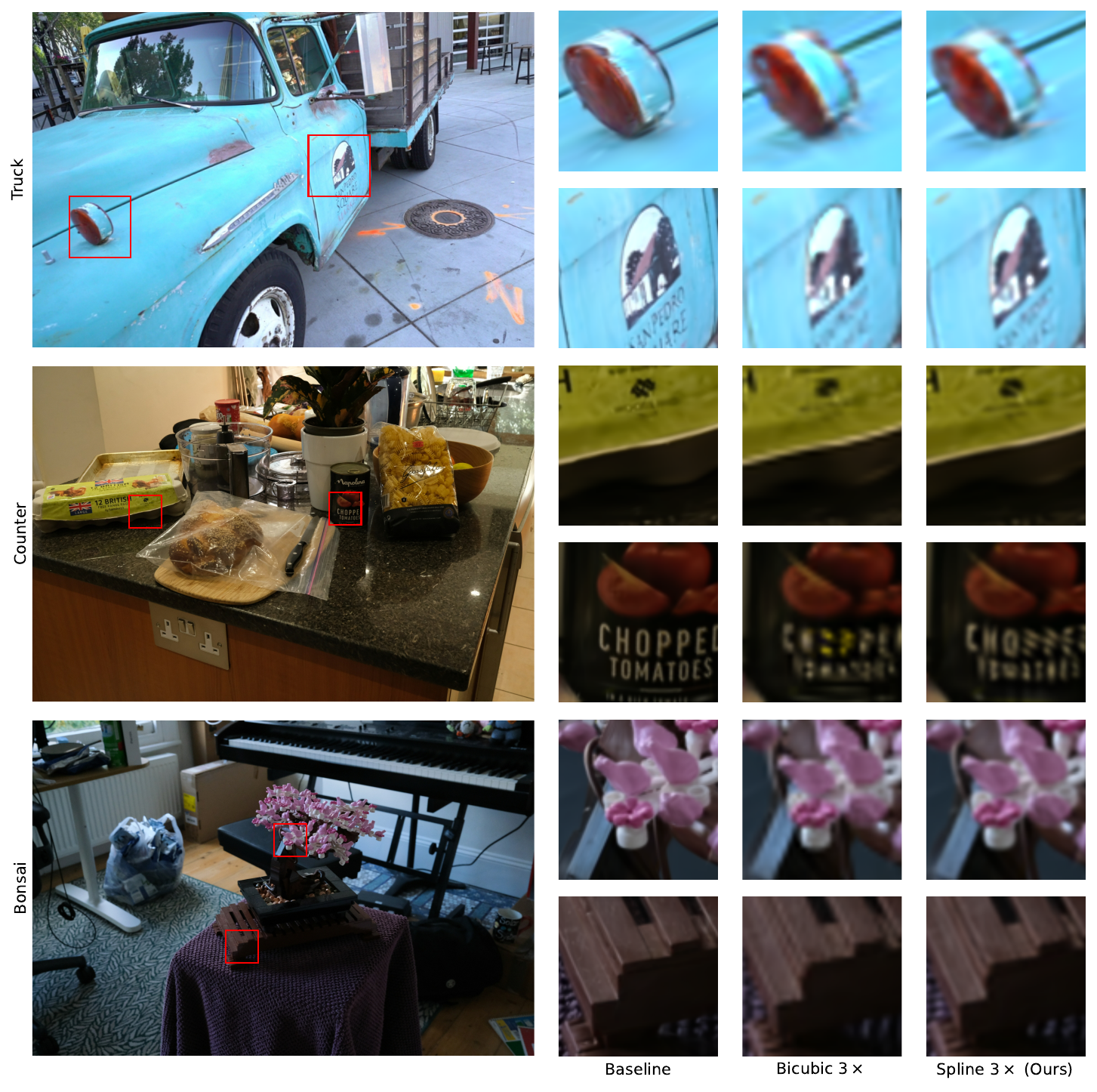}
    \caption{
    Comparison of image quality: (Left) Ground Truth, (Baseline) full resolution training with full-resolution rendering, (Bicubic) $3\times$ upscaling during training with bicubic interpolation, (Spline) $3\times$ upscaling during training with spline interpolation (ours).}
    \label{fig:comparison-crop}
\end{figure*}

We provided detailed metrics/results for all the experiments in the paper.
\cref{tab:metrics-2x,tab:metrics-4x,tab:metrics-8x} show detailed results for training with upscaling for each scene from the different datasets used in the paper.
Additionally, in \cref{fig:comparison-crop}, we provide additional visual results for the scenes not shown in the paper.


\begin{table}[h]
    \centering
    \begin{tabular}{lllll}
    \toprule
    Method & Scale & Render & Upscale & Speedup  \\
    \midrule
    3DGS & 1 & 72.4 ms & - & - \\
    \midrule
    \multirow[t]{4}{*}{Bicubic} & 2 & 24.6 ms & 2.0 ms & $\times2.7$ \greenup \\
     & 3 & 15.4 ms & 1.8 ms & $\times4.2$ \greenup \\
     & 4 & 11.8 ms & 1.8 ms & $\times5.3$ \greenup \\
     & 8 & 10.5 ms & 1.6 ms & $\times6.0$ \greenup \\
    \midrule
    \multirow[t]{2}{*}{DLSS} & 2 & 24.0 ms & 8.9 ms & $\times2.2$ \greenup \\
     & 3 & 15.1 ms & 8.0 ms & $\times3.1$ \greenup \\
    \midrule
    \multirow[t]{4}{*}{NinaSR-B1} & 2 & 23.8 ms & 604.8 ms & $\times0.1$ \reddown \\
     & 3 & 15.0 ms & 275.3 ms & $\times0.2$ \reddown \\
     & 4 & 11.7 ms & 158.9 ms & $\times0.4$ \reddown \\
     & 8 & 10.7 ms & 45.4 ms & $\times1.3$ \greenup \\
    \midrule
    \multirow[t]{4}{*}{Spline (Ours)} & 2 & 24.7 ms & 4.0 ms & $\times2.5$ \greenup \\
     & 3 & 15.6 ms & 1.8 ms & $\times4.2$ \greenup \\
     & 4 & 11.9 ms & 2.2 ms & $\times5.1$ \greenup \\
     & 8 & 10.7 ms & 1.7 ms & $\times5.9$ \greenup \\
    \bottomrule
    \end{tabular}
    \caption{Render times for different methods and upscaling factors. The target resolution is 5187x3361. The garden scene with 5 Million Gaussians is used. The cameras from the test set are used, and measurements are averaged over all images.}
    \label{tab:render_times_all}
\end{table}


\subsection{DL-based Upscalers}

Sec.~6.3 of the main manuscript discusses the results achieved via DL-based upscaling techniques. DL-based upscalers can be compared to our proposed spline-based upscaling technique with respect to upscaling quality, as in Fig.~6 of the main manuscript. Furthermore, a DL-based upscaler can be incorporated into the training pipeline for low-resolution renderings, as illustrated in Fig.~4 of the main manuscript. In this section, we elaborate on this incorporation.

We considered many pre-trained DL-based upscalers  from the torchSR project~\cite{torchsr}, of which  EDSR~\cite{edsr} is supposed to achieve the best results.
However, incorporating this network into the 3DGS training pipeline turned out to be impossible, as the model ran out of memory on the NVIDIA L4 GPU with 24GB of video memory.
Therefore, we decided to use the smaller DL-based upscale NinaSR-B1~\cite{ninasr}, which the torchSR project recommends for ``practical applications'' \cite{torchsr}.

However, when training models with low-resolution renderings upscaled to the ground truth image resolution via the pre-trained NinaSR-B1, we could not achieve high-quality results.
Firstly, even with NinaSR-B1 we were running out of memory when training with ground truth images of resolution $2594\times1681$ (called images\_2 in the dataset) for the Garden scene from Mip-NeRF~\cite{barron_mip-nerf_2022}. 
Consequently, we decided to use an upscaled resolution of $1296\times840$ pixels (images\_4).
Qualitative and quantitative results can be seen in \cref{fig:ninasr-train}.
The deep learning-based upscaler produces overly sharp images that are not consistent across different views, leading to an overall worse result.
Our spline-based approach is able to achieve significantly higher PSNR values (cf.~\cref{fig:ninasr-train}).


\begin{table*}[h]
\resizebox{\linewidth}{!}{%
\begin{tabular}{ll|SSSS|SSSS|SSSS}
\toprule
{} & {} & \multicolumn{4}{c}{3DGS} & \multicolumn{4}{c}{Bicubic} & \multicolumn{4}{c}{Spline (Ours)} \\
{Dataset} & {Scene} & {PSNR} & {SSIM} & {LPIPS} & {Duration} & {PSNR} & {SSIM} & {LPIPS} & {Duration} & {PSNR} & {SSIM} & {LPIPS} & {Duration} \\
\midrule
\multirow[c]{9}{*}{MipNeRF360} & bicycle & 24.402 & 0.733 & 0.338 & 116 & 24.434 & 0.737 & 0.340 & 141 & 24.397 & 0.734 & 0.344 & 162 \\
 & bonsai & 32.021 & 0.940 & 0.267 & 76 & 31.520 & 0.937 & 0.270 & 87 & 32.071 & 0.940 & 0.268 & 95 \\
 & counter & 28.507 & 0.912 & 0.255 & 89 & 28.374 & 0.915 & 0.254 & 99 & 28.391 & 0.915 & 0.253 & 107 \\
 & flowers & 20.871 & 0.598 & 0.414 & 119 & 20.984 & 0.604 & 0.418 & 145 & 20.955 & 0.603 & 0.420 & 165 \\
 & garden & 26.351 & 0.794 & 0.261 & 130 & 26.281 & 0.800 & 0.260 & 154 & 26.330 & 0.799 & 0.263 & 178 \\
 & kitchen & 31.277 & 0.920 & 0.186 & 90 & 31.093 & 0.922 & 0.188 & 99 & 31.022 & 0.923 & 0.185 & 109 \\
 & room & 31.549 & 0.921 & 0.269 & 82 & 31.585 & 0.923 & 0.268 & 93 & 31.657 & 0.923 & 0.266 & 101 \\
 & stump & 26.248 & 0.797 & 0.366 & 114 & 26.446 & 0.804 & 0.364 & 137 & 26.442 & 0.803 & 0.365 & 160 \\
 & treehill & 21.490 & 0.669 & 0.418 & 126 & 21.788 & 0.679 & 0.425 & 150 & 21.732 & 0.678 & 0.427 & 173 \\
 \midrule
\multirow[c]{2}{*}{Deep Blending} & drjohnson & 29.259 & 0.898 & 0.251 & 17 & 29.258 & 0.899 & 0.251 & 18 & 29.120 & 0.901 & 0.249 & 19 \\
 & playroom & 30.029 & 0.904 & 0.243 & 17 & 30.090 & 0.905 & 0.246 & 17 & 30.266 & 0.909 & 0.242 & 18 \\
 \midrule
\multirow[c]{2}{*}{Tanks and Temples} & train & 19.966 & 0.754 & 0.254 & 18 & 20.201 & 0.761 & 0.265 & 19 & 20.622 & 0.788 & 0.235 & 19 \\
 & truck & 24.219 & 0.844 & 0.185 & 16 & 24.173 & 0.837 & 0.211 & 17 & 24.652 & 0.864 & 0.177 & 17 \\
\bottomrule
\end{tabular}
}
\caption{Training with $2\times$ upscaling at full resolution compared to 3DGS with no upscaling in training. Duration is reported in minutes.}
\label{tab:metrics-2x}
\bigskip

\resizebox{\linewidth}{!}{%
\begin{tabular}{ll|SSSS|SSSS|SSSS}
\toprule
{} & {} & \multicolumn{4}{c}{3DGS} & \multicolumn{4}{c}{Bicubic} & \multicolumn{4}{c}{Spline (Ours)} \\
{Dataset} & {Scene} & {PSNR} & {SSIM} & {LPIPS} & {Duration} & {PSNR} & {SSIM} & {LPIPS} & {Duration} & {PSNR} & {SSIM} & {LPIPS} & {Duration} \\
\midrule
\multirow[c]{9}{*}{MipNeRF360} & bicycle & 24.216 & 0.708 & 0.375 & 60 & 24.222 & 0.722 & 0.365 & 95 & 24.317 & 0.729 & 0.352 & 105 \\
 & bonsai & 30.728 & 0.915 & 0.310 & 50 & 31.053 & 0.925 & 0.293 & 63 & 31.110 & 0.934 & 0.282 & 68 \\
 & counter & 27.082 & 0.875 & 0.298 & 54 & 27.906 & 0.899 & 0.291 & 68 & 28.162 & 0.909 & 0.269 & 72 \\
 & flowers & 20.630 & 0.571 & 0.440 & 63 & 21.010 & 0.598 & 0.433 & 95 & 20.984 & 0.601 & 0.428 & 106 \\
 & garden & 25.834 & 0.755 & 0.322 & 67 & 25.916 & 0.767 & 0.308 & 100 & 26.192 & 0.786 & 0.282 & 110 \\
 & kitchen & 30.194 & 0.886 & 0.246 & 56 & 30.258 & 0.900 & 0.237 & 70 & 31.262 & 0.920 & 0.199 & 73 \\
 & room & 31.003 & 0.905 & 0.293 & 52 & 30.933 & 0.909 & 0.303 & 65 & 31.349 & 0.918 & 0.283 & 68 \\
 & stump & 26.050 & 0.780 & 0.389 & 60 & 26.318 & 0.800 & 0.372 & 93 & 26.336 & 0.801 & 0.367 & 104 \\
 & treehill & 21.400 & 0.646 & 0.447 & 65 & 21.801 & 0.673 & 0.444 & 99 & 21.941 & 0.680 & 0.435 & 110 \\
 \midrule
\multirow[c]{2}{*}{Deep Blending} & drjohnson & 28.596 & 0.873 & 0.286 & 14 & 28.433 & 0.871 & 0.310 & 15 & 28.750 & 0.887 & 0.286 & 16 \\
 & playroom & 29.180 & 0.880 & 0.285 & 14 & 28.958 & 0.880 & 0.313 & 15 & 29.424 & 0.893 & 0.286 & 15 \\
 \midrule
\multirow[c]{2}{*}{Tanks and Temples} & train & 18.944 & 0.651 & 0.356 & 21 & 18.457 & 0.636 & 0.412 & 21 & 19.598 & 0.708 & 0.341 & 22 \\
 & truck & 21.684 & 0.703 & 0.325 & 17 & 22.096 & 0.719 & 0.372 & 18 & 23.002 & 0.780 & 0.305 & 17 \\
\bottomrule
\end{tabular}

}
\caption{Training with $4\times$ upscaling at full resolution compared to 3DGS with no upscaling in training. Duration is reported in minutes.}
\label{tab:metrics-4x}
\bigskip

\resizebox{\linewidth}{!}{%
\begin{tabular}{ll|SSSS|SSSS|SSSS}
\toprule
{} & {} & \multicolumn{4}{c}{3DGS} & \multicolumn{4}{c}{Bicubic} & \multicolumn{4}{c}{Spline (Ours)} \\
{Dataset} & {Scene} & {PSNR} & {SSIM} & {LPIPS} & {Duration} & {PSNR} & {SSIM} & {LPIPS} & {Duration} & {PSNR} & {SSIM} & {LPIPS} & {Duration} \\
\midrule
\multirow[c]{9}{*}{MipNeRF360} & bicycle & 22.941 & 0.601 & 0.467 & 47 & 23.008 & 0.657 & 0.479 & 88 & 23.456 & 0.683 & 0.430 & 90 \\
 & bonsai & 28.071 & 0.822 & 0.418 & 47 & 29.310 & 0.885 & 0.370 & 63 & 29.915 & 0.911 & 0.322 & 64 \\
 & counter & 25.848 & 0.774 & 0.407 & 49 & 26.451 & 0.851 & 0.394 & 67 & 27.609 & 0.882 & 0.333 & 66 \\
 & flowers & 19.658 & 0.473 & 0.500 & 48 & 20.641 & 0.563 & 0.498 & 90 & 20.697 & 0.578 & 0.475 & 92 \\
 & garden & 23.861 & 0.621 & 0.438 & 50 & 24.276 & 0.665 & 0.445 & 94 & 25.064 & 0.712 & 0.383 & 96 \\
 & kitchen & 27.245 & 0.753 & 0.413 & 50 & 27.524 & 0.802 & 0.385 & 68 & 29.168 & 0.871 & 0.287 & 68 \\
 & room & 29.320 & 0.854 & 0.366 & 45 & 29.427 & 0.878 & 0.378 & 62 & 29.785 & 0.895 & 0.336 & 62 \\
 & stump & 24.567 & 0.666 & 0.488 & 46 & 25.496 & 0.764 & 0.455 & 88 & 25.934 & 0.786 & 0.413 & 90 \\
 & treehill & 20.532 & 0.561 & 0.516 & 49 & 21.652 & 0.644 & 0.515 & 91 & 21.907 & 0.666 & 0.488 & 92 \\
\bottomrule
\end{tabular}

}
\caption{Training with $8\times$ upscaling at full resolution compared to 3DGS with no upscaling in training. Duration is reported in minutes.}
\label{tab:metrics-8x}
\end{table*}

\cleardoublepage
\clearpage

\subsection{Spline Image Interpolation}

A bicubic spline can be parameterized with a third-order polynomial with the coefficients $A \in \mathbb{R}^{4\times4} $:
\begin{equation}
p(x,y) = \sum_{i=0}^{3}\sum_{j=0}^{3} a_{ij}x^i y^j
\end{equation}

The partial derivatives of the spline are given by:
\begin{align}
\frac{\partial p(x,y)}{\partial x} &= \sum_{i=1}^{3}\sum_{j=0}^{3} i a_{ij}x^{i-1} y^j \\
\frac{\partial p(x,y)}{\partial y} &= \sum_{i=0}^{3}\sum_{j=1}^{3} j a_{ij}x^i y^{j-1} \\
\frac{\partial^2 p(x,y)}{\partial x \partial y} &= \sum_{i=1}^{3}\sum_{j=1}^{3} i j a_{ij}x^{i-1} y^{j-1}
\end{align}

The corner values $f(0,0), f(1,0), f(0,1), f(1,1)$ and their derivatives, e.g., $\frac{\partial f(0,0)}{\partial x},\frac{\partial f(0,0)}{\partial y},\frac{\partial f(0,0)}{\partial x \partial y}$ are known.
The value of the spline and its derivatives must be equal to $f(x,y)$ at the corner points.

The problem of calculating the coefficients $A$ can be formulated as a linear problem:
\begin{align}
F &= CAC^T \\
A &=  C^{-1}F (C^T)^{-1}\\
F &= \begin{bmatrix}
f(0,0)&f(0,1)&f_y(0,0)&f_y(0,1)\\
f(1,0)&f(1,1)&f_y(1,0)&f_y(1,1)\\
f_x(0,0)&f_x(0,1)&f_{xy}(0,0)&f_{xy}(0,1)\\
f_x(1,0)&f_x(1,1)&f_{xy}(1,0)&f_{xy}(1,1)\\
\end{bmatrix}.
\end{align}

With $C$ being the coefficients of the cubic spline at the respective points:
\begin{align}
C &= \begin{bmatrix}
1 & 0 & 0 & 0 \\
1 & 1 & 1 & 1 \\
0 & 1 & 0 & 0 \\
0 & 1 & 2 & 3 \\
\end{bmatrix}
\end{align}

The coefficients are derived from a cubic function and its derivative at $0$ and $1$:

\begin{align}
f(x) &=   a_0   &+& a_1 x &+& a_2 x^2 &+& a_3 x^3 \\
f_x(x) &=  &+& a_1   &+& 2 a_2 x &+& 3 a_3 x^2 \\
f(0) &=   1a_0   &+& 0     &+& 0       &+& 0 \\
f(1) &=   1a_0   &+& 1a_1   &+& 1a_2     &+& 1a_3 \\
f_x(0) &= 0     &+& 1a_1   &+& 0       &+& 0  \\
f_x(1) &= 0     &+& 1a_1   &+& 2a_2    &+& 3 a_3
\end{align}

We refer to ~\cite{RUSSELL1995129} for a more detailed explanation.

\section{3DGS Gradient Computation}

In the following, we describe how the screen space gradients are computed in the forward pass, and we dicsuss the backward pass required for training with differentiable image upscaling.

\subsection{Forward pass}

With a 3DGS model comprising \(N\) Gaussians, the image \(I(x,y)\) is rendered as follows: 

\begin{align}
    I(x,y) &= \sum_{i=1}^{N} T_i(x,y) \alpha_i(x,y) c_i \\
    \alpha_i(x,y) &= \sigma_i \exp(g_i(x,y)) \\
    g_i(x,y) &= -d_i\Sigma_i d_i^T \\
    d_i &= \begin{bmatrix}x-\mu_x&y-\mu_y\end{bmatrix} \\
    T_i(x,y) &= \prod_{j=1}^{i-1}(1-\alpha_j(x,y))
\end{align}

Here, \(c_i\) is the color of the i-th Gaussian and \(\sigma_i\) its
opacity. \(\Sigma_i \in \mathbb{R}^{2\times2}\) is the 2D covariance
matrix calculated with EWA~\cite{zwicker_ewa_2001} splatting for each Gaussian.

The image computation can be reformulated using front-to-back
\(\alpha\)-blending:

\begin{align}
    I(x,y) &= B_N \\
    B_i &= B_{i-1} + (1-A_{i-1})\alpha_i c_i \\
    A_i &= A_{i-1} + \alpha_i(1-A_{i-1}) \\
    B_0 &= 0\\
    A_0 &= 0
\end{align}

Using this formulation, the partial derivatives with respect to \(x\) and
\(y\) can be computed as 

\begin{align}
\frac{\partial I(x,y)}{\partial x} &= \frac{\partial B_N}{\partial x} \\
\frac{\partial B_i}{\partial x} &= \frac{\partial B_{i-1}}{\partial x} + c_i ((1-A_{i-1})\frac{\partial \alpha_i}{\partial x} - \frac{\partial A_{i-1}}{\partial x}\alpha_i ) \\
\frac{\partial B_i}{\partial y} &= \frac{\partial B_{i-1}}{\partial y} + c_i ((1-A_{i-1})\frac{\partial \alpha_i}{\partial y} - \frac{\partial A_{i-1}}{\partial y}\alpha_i ) \\
\frac{\partial^2 B_i}{\partial x \partial y} &= \frac{\partial^2 B_{i-1}}{\partial x \partial y} + c_i ( (1-A_{i-1}) \frac{\partial^2 \alpha_i}{\partial x\partial y} - \frac{\partial A_{i-1}}{\partial y} \frac{\partial \alpha_i}{\partial x} \\ & {}  - \frac{\partial^2 A_{i-1}}{\partial x\partial y} \alpha_i - \frac{\partial A_{i-1}}{\partial x} \frac{\partial \alpha_i}{\partial y} )
\end{align}

The partial derivatives of \(A_i\) are 
\begin{align}
    \frac{\partial A_i}{\partial x} &= \frac{\partial A_{i-1}}{\partial x}  (1-\alpha_i) + (1-A_{i-1}) \frac{\partial \alpha_i}{\partial x} \\
    \frac{\partial A_i}{\partial y} &= \frac{\partial A_{i-1}}{\partial y}  (1-\alpha_i) + (1-A_{i-1}) \frac{\partial \alpha_i}{\partial y}\\
    \frac{\partial^2 A_i}{\partial x \partial y} &= 
    \begin{aligned}[t] \frac{\partial^2 A_{i-1}}{\partial x \partial y} (1-\alpha_i) +(1-A_{i-1}) \frac{\partial^2 \alpha_i}{\partial x\partial y}\\
         {}- \frac{\partial A_{i-1}}{\partial x}  \frac{\partial \alpha_i}{\partial y} - \frac{\partial A_{i-1}}{\partial y}\frac{\partial \alpha_i}{\partial x} 
    \end{aligned}
\end{align}

With this formulation, the computation of the image gradients
\(\frac{\partial I(x,y)}{\partial x}\),
\(\frac{\partial I(x,y)}{\partial y}\),
\(\frac{\partial^2 I(x,y)}{\partial x \partial y}\) can be integrated
into the rendering loop of 3DGS without much overhead.

\hypertarget{backward-pass}{%
\subsection{Backward Pass}\label{backward-pass}}

For the backward pass, the partial derivatives of the image gradients
with respect to the 2D Gaussian parameters must be computed. As in 3DGS,
we use back-to-front blending in the backward pass:
\begin{align}
    I(x,y) &= \hat{B}_1 \\
    \hat{B}_i &= (1-\alpha_i) \hat{B}_{i+1} + \alpha_i c_i \\
    \hat{B}_{N+1} &= 0
\end{align}

The derivatives with respect to the screen space positions are

\begin{align}
\frac{\partial I(x,y)}{\partial x} &= \frac{\partial \hat{B}_1}{\partial x} \\
\frac{\partial \hat{B}_i}{\partial x} &= (1-\alpha_i) \frac{\partial \hat{B}_{i+1}}{\partial x} + \frac{\partial \alpha_i}{\partial x} (c_i - \hat{B}_{i+1}) \\
\frac{\partial \hat{B}_i}{\partial y} &= (1-\alpha_i) \frac{\partial \hat{B}_{i+1}}{\partial y} + \frac{\partial \alpha_i}{\partial x} (c_i - \hat{B}_{i+1}) \\
\frac{\partial^2 \hat{B}_i}{\partial x \partial y} &= \begin{aligned}[t](1-\alpha_i) \frac{\partial^2 \hat{B}_{i+1}}{\partial x\partial y} + \frac{\partial^2 \alpha_i}{\partial x \partial y} (c_i - \hat{B}_{i+1})
\\ - \frac{\partial \alpha_i}{\partial y}\frac{\partial \hat{B}_{i+1}}{\partial x}  - \frac{\partial \alpha_i}{\partial x} \frac{\partial \hat{B}_{i+1}}{\partial y} \end{aligned}
\end{align}

The gradients for \(\sigma,\Sigma,\mu_x,\mu_y\) are computed analogously. We demonstrate the computation of \(\sigma\) as an example. The chain rule is applied to compute the partial derivatives of \(\sigma_k\):
\begin{align}
\frac{\partial I(x,y)}{\partial \sigma_k} &= \frac{\partial I(x,y)}{\partial \alpha_k} \frac{\partial \alpha_k}{\partial \sigma_k} \\
\frac{\partial^2 I(x,y)}{\partial \sigma_k \partial x} &=  \frac{\partial^2 I(x,y)}{\partial \alpha_k \partial x} \frac{\partial \alpha_k}{\partial \sigma_k} +  \frac{\partial I(x,y)}{\partial \alpha_k} \frac{\partial^2 \alpha_k}{\partial \sigma_k \partial x}\\
\frac{\partial^2 I(x,y)}{\partial \sigma_k \partial y} &=  \frac{\partial^2 I(x,y)}{\partial \alpha_k \partial y} \frac{\partial \alpha_k}{\partial \sigma_k} +  \frac{\partial I(x,y)}{\partial \alpha_k} \frac{\partial^2 \alpha_k}{\partial \sigma_k \partial y}\\
\frac{\partial^3 I(x,y)}{\partial \sigma_k \partial y \partial x} &= \begin{aligned}[t] \frac{\partial^3 I(x,y)}{\partial \alpha_k \partial y\partial x} \frac{\partial \alpha_k}{\partial \sigma_k} + \frac{\partial^2 I(x,y)}{\partial \alpha_k \partial y} \frac{\partial^2 \alpha_k}{\partial \sigma_k \partial x} \\ {}+  \frac{\partial^2 I(x,y)}{\partial \alpha_k \partial x} \frac{\partial^2 \alpha_k}{\partial \sigma_k \partial x} + \frac{\partial I(x,y)}{\partial \alpha_k} \frac{\partial^3 \alpha_k}{\partial \sigma_k \partial x \partial y}
\end{aligned}
\end{align}

The derivative of \(I(x,y)\) with respect to \(\alpha_a\) is given by

\begin{align}
\frac{\partial I(x,y)}{\partial \alpha_k}  &= \frac{\partial \hat{B}_1}{\partial \alpha_k} \\
 \frac{\partial \hat{B}_i}{\partial \alpha_k} &= \delta_{i< k} (\frac{\partial \hat{B}_{i+1}}{\partial \alpha_k} (1-\alpha_i) ) + \delta_{i = k} (c_i-\hat{B}_{i+1}) \\ &= \prod_{j=1}^{k-1}(1-\alpha_j) (c_k-\hat{B}_{k+1})\\
 &= (1-A_k) (c_k-\hat{B}_{k+1})
\end{align}

Here, $\delta_{i,j}$ is the Kronecker delta, which equals one if the condition in the subscript is true and zero otherwise.
Following this, we calculate the derivative with respect to the screen positions
\(x,y\):

\begin{align}
\frac{\partial^2 \hat{B}_i}{\partial \alpha_k \partial x} &= -\frac{\partial A_k}{\partial x} (c_k-\hat{B}_{k+1}) - (1-A_k) \frac{\partial \hat{B}_{k+1}}{\partial x} \\
\frac{\partial^2 \hat{B}_i}{\partial \alpha_k \partial y} &= -\frac{\partial A_k}{\partial y} (c_k-\hat{B}_{k+1}) - (1-A_k) \frac{\partial \hat{B}_{k+1}}{\partial y} \\
\frac{\partial^3 \hat{B}_i}{\partial \alpha_k \partial x \partial y} &= \begin{aligned}[t] -\frac{\partial^2 A_k}{\partial x \partial y} (c_k-\hat{B}_{k+1})- (1-A_k) \frac{\partial^2 \hat{B}_{k+1}}{\partial x \partial y} \\ + \frac{\partial A_k}{\partial x} \frac{\partial \hat{B}_{k+1}}{\partial y} +\frac{\partial A_k}{\partial y} \frac{\partial \hat{B}_{k+1}}{\partial x} 
\end{aligned}
\end{align}

\paragraph{Splat Color Gradients}
The calculation of the partial derivatives of a Gaussian's color is straightforward, i.e., 
\begin{align}
 \frac{\partial \hat{B}_i}{\partial c_k} &= \delta_{i< k} \frac{\partial \hat{B}_{i+1}}{\partial \alpha_k} (1-\alpha_i) + \delta_{i = k} \alpha_i \\&= \prod_{j=1}^{k-1}(1-\alpha_j) \alpha_k = (1-A_k) \alpha_k \label{eq:color_grad}
\end{align}

Differentiating with respect to the screen space position yields the following partial derivatives:

\begin{align}
 \frac{\partial \hat{B}_i}{\partial c_k \partial x} &= (1-A_k) \frac{\partial \alpha_k}{\partial x} - \frac{\partial A_k}{\partial x} \alpha_k \\
 \frac{\partial \hat{B}_i}{\partial c_k \partial y} &= (1-A_k) \frac{\partial \alpha_k}{\partial y} - \frac{\partial A_k}{\partial y} \alpha_k\\
 \frac{\partial^3 \hat{B}_i}{\partial c_k \partial x \partial y} &=  \begin{aligned}[t](1-A_k) \frac{\partial^2 \alpha_k}{\partial x \partial y} - \frac{\partial^2 A_k}{\partial x \partial y} \alpha_k \\- \frac{\partial A_k}{\partial y}\frac{\partial \alpha_k}{\partial x} - \frac{\partial A_k}{\partial x}\frac{\partial \alpha_k}{\partial y}\end{aligned}
\end{align}

\paragraph{Inversion Trick}
In the forward pass, we store
\(A_N,\frac{\partial A_N}{\partial x},\frac{\partial A_N}{\partial y},\frac{\partial^2 A_N}{\partial x \partial y}\)
for each pixel. To calculate \(A_i\) and its derivatives in the backward
pass we use the Inversion trick~\cite{weiss_differentiable_2022}:
\begin{align}
    A_{i-1} &= \frac{A_i  \alpha_i}{1-\alpha_i} \\
    \frac{\partial A_{i-1}}{\partial x} &= \frac{1}{1-\alpha_i}(\frac{\partial A_i}{\partial x} - (1-A_{i-1})\frac{\partial \alpha_i}{\partial x}) \\
    \frac{\partial A_{i-1}}{\partial y} &= \frac{1}{1-\alpha_i}(\frac{\partial A_i}{\partial y} - (1-A_{i-1})\frac{\partial \alpha_i}{\partial y}) \\
    \frac{\partial A_{i-1}}{\partial x \partial y} &=  \begin{aligned}[t]\frac{1}{1-\alpha_i}(\frac{\partial^2 A_{i}}{\partial x \partial y} - (1-A_{i-1}) \frac{\partial^2 \alpha_i}{\partial x\partial y} \\ + \frac{\partial A_{i-1}}{\partial x}  \frac{\partial \alpha_i}{\partial y} - \frac{\partial A_{i-1}}{\partial y}\frac{\partial \alpha_i}{\partial x}) 
    \end{aligned}
\end{align}

%% file: main.bbl
\begin{thebibliography}{35}
\providecommand{\natexlab}[1]{#1}
\providecommand{\url}[1]{\texttt{#1}}
\expandafter\ifx\csname urlstyle\endcsname\relax
  \providecommand{\doi}[1]{doi: #1}\else
  \providecommand{\doi}{doi: \begingroup \urlstyle{rm}\Url}\fi

\bibitem[{AMD}(2021)]{amd_fsr}
{AMD}.
\newblock {AMD FidelityFX Super Resolution (FSR)}, 2021.
\newblock
  \url{https://www.amd.com/en/technologies/fidelityfx-super-resolution}.

\bibitem[Bangaru et~al.(2023)Bangaru, Wu, Li, Munkberg, Bernstein,
  Ragan-Kelley, Durand, Lefohn, and He]{bangaru2023slangd}
Sai Bangaru, Lifan Wu, Tzu-Mao Li, Jacob Munkberg, Gilbert Bernstein, Jonathan
  Ragan-Kelley, Fredo Durand, Aaron Lefohn, and Yong He.
\newblock Slang.d: Fast, modular and differentiable shader programming.
\newblock \emph{ACM Transactions on Graphics (SIGGRAPH Asia)}, 42\penalty0
  (6):\penalty0 1--28, 2023.

\bibitem[Barron et~al.(2022)Barron, Mildenhall, Verbin, Srinivasan, and
  Hedman]{barron_mip-nerf_2022}
Jonathan~T. Barron, Ben Mildenhall, Dor Verbin, Pratul~P. Srinivasan, and Peter
  Hedman.
\newblock Mip-{NeRF} 360: {Unbounded} {Anti}-{Aliased} {Neural} {Radiance}
  {Fields}.
\newblock In \emph{2022 {IEEE}/{CVF} {Conference} on {Computer} {Vision} and
  {Pattern} {Recognition} ({CVPR})}, pages 5460--5469, New Orleans, LA, USA,
  2022. IEEE.

\bibitem[Chen et~al.(2024)Chen, Chen, Qu, Wang, Liu, Chen, and
  Chung]{chen2024linear}
Haodong Chen, Runnan Chen, Qiang Qu, Zhaoqing Wang, Tongliang Liu, Xiaoming
  Chen, and Yuk~Ying Chung.
\newblock Beyond gaussians: Fast and high-fidelity 3d splatting with linear
  kernels, 2024.

\bibitem[Dong et~al.(2016{\natexlab{a}})Dong, Loy, He, and Tang]{dong2015image}
Chao Dong, Chen~Change Loy, Kaiming He, and Xiaoou Tang.
\newblock Image super-resolution using deep convolutional networks.
\newblock \emph{IEEE Transactions on Pattern Analysis and Machine
  Intelligence}, 38\penalty0 (2):\penalty0 295--307, 2016{\natexlab{a}}.

\bibitem[Dong et~al.(2016{\natexlab{b}})Dong, Loy, He, and
  Tang]{dong2016accelerating}
Chao Dong, Chen~Change Loy, Kaiming He, and Xiaoou Tang.
\newblock Accelerating the super-resolution convolutional neural network.
\newblock In \emph{European Conference on Computer Vision (ECCV)}, pages
  391--407. Springer, 2016{\natexlab{b}}.

\bibitem[Feng et~al.(2024)Feng, He, Wang, Yang, Li, Chen, Kuang, Ding, Fan, and
  Yu]{feng2024srgs}
Xiang Feng, Yongbo He, Yubo Wang, Yan Yang, Wen Li, Yifei Chen, Zhenzhong
  Kuang, Jiajun Ding, Jianping Fan, and Jun Yu.
\newblock {SRGS}: Super-resolution 3d gaussian splatting.
\newblock \emph{arXiv preprint arXiv:2404.10318}, 2024.

\bibitem[Getreuer(2011)]{getreuter_2011}
Pascal Getreuer.
\newblock {Linear Methods for Image Interpolation}.
\newblock \emph{{Image Processing On Line}}, 1:\penalty0 238--259, 2011.
\newblock \url{https://doi.org/10.5201/ipol.2011.g_lmii}.

\bibitem[Gouvine(2021{\natexlab{a}})]{ninasr}
Gabriel Gouvine.
\newblock {NinaSR}: Efficient small and large convnets for super-resolution.
\newblock \url{https://github.com/Coloquinte/torchSR/blob/main/doc/NinaSR.md},
  2021{\natexlab{a}}.

\bibitem[Gouvine(2021{\natexlab{b}})]{torchsr}
Gabriel Gouvine.
\newblock Super-resolution networks for {PyTorch}.
\newblock \url{https://github.com/Coloquinte/torchSR}, 2021{\natexlab{b}}.

\bibitem[Hedman et~al.(2018)Hedman, Philip, Price, Frahm, Drettakis, and
  Brostow]{hedman_deep_2018}
Peter Hedman, Julien Philip, True Price, Jan-Michael Frahm, George Drettakis,
  and Gabriel Brostow.
\newblock Deep {Blending} for {Free}-viewpoint {Image}-based {Rendering}.
\newblock \emph{ACM Transactions on Graphics (Proc. SIGGRAPH Asia)},
  37\penalty0 (6):\penalty0 257:1--257:15, 2018.
\newblock Publisher: ACM.

\bibitem[Hu et~al.(2024)Hu, Xia, Chen, Yang, and
  Zhang]{hu2024gaussiansrhighfidelity2d}
Jintong Hu, Bin Xia, Bin Chen, Wenming Yang, and Lei Zhang.
\newblock Gaussiansr: High fidelity 2d gaussian splatting for arbitrary-scale
  image super-resolution, 2024.

\bibitem[Huang et~al.(2023)Huang, Li, Hu, Chen, and Wang]{huang2023refsrnerf}
Xudong Huang, Wei Li, Jie Hu, Hanting Chen, and Yunhe Wang.
\newblock {RefSR-NeRF}: Towards high fidelity and super resolution view
  synthesis.
\newblock In \emph{Proceedings of the IEEE/CVF Conference on Computer Vision
  and Pattern Recognition (CVPR)}, pages 8244--8253, 2023.

\bibitem[Kerbl et~al.(2023)Kerbl, Kopanas, Leimkuehler, and
  Drettakis]{kerbl_3d_2023}
Bernhard Kerbl, Georgios Kopanas, Thomas Leimkuehler, and George Drettakis.
\newblock {3D} {Gaussian} {Splatting} for {Real}-{Time} {Radiance} {Field}
  {Rendering}.
\newblock \emph{ACM Trans. Graph.}, 42\penalty0 (4), 2023.
\newblock Place: New York, NY, USA Publisher: Association for Computing
  Machinery.

\bibitem[Knapitsch et~al.(2017)Knapitsch, Park, Zhou, and
  Koltun]{knapitsch_tanks_2017}
Arno Knapitsch, Jaesik Park, Qian-Yi Zhou, and Vladlen Koltun.
\newblock Tanks and temples: {Benchmarking} large-scale scene reconstruction.
\newblock \emph{ACM Transactions on Graphics (ToG)}, 36\penalty0 (4):\penalty0
  1--13, 2017.
\newblock Publisher: ACM New York, NY, USA.

\bibitem[Lai et~al.(2017)Lai, Huang, Ahuja, and Yang]{lai2017deep}
Wei-Sheng Lai, Jia-Bin Huang, Narendra Ahuja, and Ming-Hsuan Yang.
\newblock Deep laplacian pyramid networks for fast and accurate
  super-resolution.
\newblock In \emph{Proceedings of the IEEE Conference on Computer Vision and
  Pattern Recognition (CVPR)}, pages 624--632, 2017.

\bibitem[Ledig et~al.(2017)Ledig, Theis, Husz{\'a}r, Caballero, Cunningham,
  Acosta, Aitken, Tejani, Totz, Wang, and Shi]{ledig2017photo}
Christian Ledig, Lucas Theis, Ferenc Husz{\'a}r, Jose Caballero, Andrew
  Cunningham, Alejandro Acosta, Andrew Aitken, Alykhan Tejani, Johannes Totz,
  Zehan Wang, and Wenzhe Shi.
\newblock Photo-realistic single image super-resolution using a generative
  adversarial network.
\newblock In \emph{Proceedings of the IEEE Conference on Computer Vision and
  Pattern Recognition (CVPR)}, pages 4681--4690, 2017.

\bibitem[Liang et~al.(2021)Liang, Cao, Sun, Zhang, Gool, and Timofte]{SwinIR}
Jingyun Liang, Jiezhang Cao, Guolei Sun, Kai Zhang, Luc~Van Gool, and Radu
  Timofte.
\newblock Swinir: Image restoration using swin transformer, 2021.

\bibitem[Lim et~al.(2017{\natexlab{a}})Lim, Son, Kim, Nah, and Lee]{edsr}
Bee Lim, Sanghyun Son, Heewon Kim, Seungjun Nah, and Kyoung~Mu Lee.
\newblock Enhanced deep residual networks for single image super-resolution,
  2017{\natexlab{a}}.

\bibitem[Lim et~al.(2017{\natexlab{b}})Lim, Son, Kim, Nah, and
  Lee]{lim2017edsr}
Bee Lim, Sanghyun Son, Heewon Kim, Seungjun Nah, and Kyoung~Mu Lee.
\newblock {EDSR}: Enhanced deep residual networks for single image
  super-resolution.
\newblock In \emph{Proceedings of the IEEE Conference on Computer Vision and
  Pattern Recognition (CVPR) Workshops}, pages 1132--1140, 2017{\natexlab{b}}.

\bibitem[Mildenhall et~al.(2021)Mildenhall, Srinivasan, Tancik, Barron,
  Ramamoorthi, and Ng]{mildenhall_nerf_2021}
Ben Mildenhall, Pratul~P Srinivasan, Matthew Tancik, Jonathan~T Barron, Ravi
  Ramamoorthi, and Ren Ng.
\newblock Nerf: {Representing} scenes as neural radiance fields for view
  synthesis.
\newblock \emph{Communications of the ACM}, 65\penalty0 (1):\penalty0 99--106,
  2021.
\newblock Publisher: ACM New York, NY, USA.

\bibitem[Moenne-Loccoz et~al.(2024)Moenne-Loccoz, Mirzaei, Perel, de~Lutio,
  Esturo, State, Fidler, Sharp, and Gojcic]{3dgrt2024}
Nicolas Moenne-Loccoz, Ashkan Mirzaei, Or Perel, Riccardo de Lutio,
  Janick~Martinez Esturo, Gavriel State, Sanja Fidler, Nicholas Sharp, and Zan
  Gojcic.
\newblock 3d gaussian ray tracing: Fast tracing of particle scenes.
\newblock \emph{ACM Transactions on Graphics and SIGGRAPH Asia}, 2024.

\bibitem[Niedermayr et~al.(2024{\natexlab{a}})Niedermayr, Neuhauser, Petkov,
  Engel, and Westermann]{niedermayr24cinematic}
Simon Niedermayr, Christoph Neuhauser, Kaloian Petkov, Klaus Engel, and
  R\"{u}diger Westermann.
\newblock {Application of 3D Gaussian Splatting for Cinematic Anatomy on
  Consumer Class Devices}.
\newblock In \emph{Vision, Modeling, and Visualization}. The Eurographics
  Association, 2024{\natexlab{a}}.

\bibitem[Niedermayr et~al.(2024{\natexlab{b}})Niedermayr, Stumpfegger, and
  Westermann]{Niedermayr_2024_CVPR}
Simon Niedermayr, Josef Stumpfegger, and R\"{u}diger Westermann.
\newblock Compressed 3d gaussian splatting for accelerated novel view
  synthesis.
\newblock In \emph{Proceedings of the IEEE/CVF Conference on Computer Vision
  and Pattern Recognition (CVPR)}, pages 10349--10358, 2024{\natexlab{b}}.

\bibitem[{NVIDIA}(2018)]{nvidia_dlss}
{NVIDIA}.
\newblock {NVIDIA} deep learning super sampling ({DLSS}), 2018.
\newblock \url{https://www.nvidia.com/en-us/geforce/technologies/dlss/}.

\bibitem[Paszke et~al.(2017)Paszke, Gross, Chintala, Chanan, Yang, DeVito, Lin,
  Desmaison, Antiga, and Lerer]{paszke2017automatic}
Adam Paszke, Sam Gross, Soumith Chintala, Gregory Chanan, Edward Yang, Zachary
  DeVito, Zeming Lin, Alban Desmaison, Luca Antiga, and Adam Lerer.
\newblock Automatic differentiation in pytorch.
\newblock In \emph{NIPS-W}, 2017.

\bibitem[Russell(1995)]{RUSSELL1995129}
William~S Russell.
\newblock Polynomial interpolation schemes for internal derivative
  distributions on structured grids.
\newblock \emph{Applied Numerical Mathematics}, 17\penalty0 (2):\penalty0
  129--171, 1995.

\bibitem[{Saswat Mallick and Rahul Goel} et~al.(2024){Saswat Mallick and Rahul
  Goel}, Kerbl, Vicente~Carrasco, Steinberger, and De~La~Torre]{taming3dgs}
{Saswat Mallick and Rahul Goel}, Bernhard Kerbl, Francisco Vicente~Carrasco,
  Markus Steinberger, and Fernando De~La~Torre.
\newblock Taming 3dgs: High-quality radiance fields with limited resources.
\newblock In \emph{SIGGRAPH Asia 2024 Conference Papers}, 2024.

\bibitem[Wang et~al.(2022)Wang, Wu, Guo, Zhang, Tai, and Hu]{wang2022nerfsr}
Chen Wang, Xian Wu, Yuan-Chen Guo, Song-Hai Zhang, Yu-Wing Tai, and Shi-Min Hu.
\newblock {NeRF-SR}: High-quality neural radiance fields using supersampling.
\newblock In \emph{Proceedings of the 30th ACM International Conference on
  Multimedia (ACM MM)}, pages 334--344, 2022.

\bibitem[Wang et~al.(2018)Wang, Yu, Wu, Gu, Liu, Dong, Loy, Qiao, and
  Tang]{wang2018esrgan}
Xintao Wang, Ke Yu, Shixiang Wu, Jinjin Gu, Yihao Liu, Chao Dong, Chen~Change
  Loy, Yu Qiao, and Xiaoou Tang.
\newblock {ESRGAN}: Enhanced super-resolution generative adversarial networks.
\newblock In \emph{Proceedings of the European Conference on Computer Vision
  (ECCV) Workshops}, pages 63--79, 2018.

\bibitem[Wang et~al.(2021)Wang, Xie, Dong, and Shan]{wang2021realesrgan}
Xintao Wang, Liangbin Xie, Chao Dong, and Ying Shan.
\newblock Real-esrgan: Training real-world blind super-resolution with pure
  synthetic data.
\newblock \emph{arXiv preprint arXiv:2107.10833}, 2021.

\bibitem[Weiss and Westermann(2022)]{weiss_differentiable_2022}
Sebastian Weiss and R\"{u}diger Westermann.
\newblock Differentiable {Direct} {Volume} {Rendering}.
\newblock In \emph{{IEEE} {Transactions} on {Visualization} and {Computer}
  {Graphics}}, pages 562--572, 2022.
\newblock Issue: 1.

\bibitem[Xie et~al.(2024)Xie, Wang, Zhu, and Pan]{xie2024supergs}
Shiyun Xie, Zhiru Wang, Yinghao Zhu, and Chengwei Pan.
\newblock {SuperGS}: Super-resolution 3d gaussian splatting via latent feature
  field and gradient-guided splitting.
\newblock \emph{arXiv preprint arXiv:2410.02571}, 2024.

\bibitem[Yu et~al.(2024)Yu, Chen, Huang, Sattler, and
  Geiger]{Yu2024MipSplatting}
Zehao Yu, Anpei Chen, Binbin Huang, Torsten Sattler, and Andreas Geiger.
\newblock Mip-splatting: Alias-free 3d gaussian splatting.
\newblock In \emph{Proceedings of the IEEE/CVF Conference on Computer Vision
  and Pattern Recognition (CVPR)}, pages 19447--19456, 2024.

\bibitem[Zwicker et~al.(2001)Zwicker, Pfister, Van~Baar, and
  Gross]{zwicker_ewa_2001}
M. Zwicker, H. Pfister, J. Van~Baar, and M. Gross.
\newblock {EWA} volume splatting.
\newblock In \emph{Proceedings {Visualization}, 2001. {VIS} '01.}, pages
  29--538, San Diego, CA, USA, 2001. IEEE.

\end{thebibliography}
